\documentclass[11pt,a4paper]{article}
\usepackage{authblk} 

\usepackage{marginnote} 
\usepackage{algorithm} 
\usepackage[noend]{algpseudocode}  

\usepackage{times,latexsym}
\usepackage{url}
\usepackage[T1]{fontenc}
\usepackage{amsfonts}

\usepackage[utf8]{inputenc}

\usepackage{microtype}

\usepackage{inconsolata}

\usepackage{graphicx}
\usepackage{subcaption}

\usepackage{bytefield}
\usepackage{tabularx}
\usepackage{longtable}  

\usepackage{booktabs}   
\usepackage{array}
\usepackage{caption}
\usepackage{makecell} 
\usepackage{relsize}
\newcommand{\FactorName}[1]{\textsmaller[0.9]{\textsc{#1}}}

\newcommand{\amRemove}[1]
{\textcolor{red}{}}

\newcommand{\rtRemove}[1]
{\textcolor{red}{}}
\usepackage{amsmath}
\newcommand{\LambdaMat}{\boldsymbol{\Lambda}} 
\newcommand{\ThetaMat}{\boldsymbol{\Theta}} 
\newcommand{\epsilonMat}{\boldsymbol{\epsilon}} 
\newcommand{\PerformanceMat}{\boldsymbol{\Pi}}

\usepackage{amsmath,amssymb,bm}    
\usepackage{enumitem}              
\usepackage{hyperref}              
\usepackage[nameinlink]{cleveref}  
\crefname{section}{\S}{\S\S}
\Crefname{section}{\S}{\S\S}
\crefname{subsection}{\S}{\S\S}
\Crefname{subsection}{\S}{\S\S}

\usepackage[acceptedWithA]{tacl2021v1}

\usepackage{xspace,mfirstuc,tabulary}

\newif\iftaclinstructions
\taclinstructionsfalse 
\iftaclinstructions
\renewcommand{\confidential}{}
\renewcommand{\anonsubtext}{(No author info supplied here, for consistency with
TACL-submission anonymization requirements)}
\newcommand{\instr}
\fi

\iftaclpubformat 

\else

\fi

\title{From Benchmarks to Skills: \\Low-Rank Factors for LLM Evaluation}

\author{
\textbf{Aviya Maimon}$^{1,2}$ \quad
\textbf{Amir David Nisan Cohen}$^{2}$ \quad
\textbf{Gal Vishne}$^{3}$ \\
\textbf{Shauli Ravfogel}$^{4}$ \quad
\textbf{Reut Tsarfaty}$^{1}$ \\
\\
$^{1}$Bar-Ilan University \quad
$^{2}$OriginAI \quad
$^{3}$Data Science Institute, Columbia University \\
$^{4}$Center for Data Science, New York University \\
\\
\texttt{\{aviyamn, amirdnc, gal.vishne, shauli.ravfogel, reut.tsarfaty\}@gmail.com}
}

\date{}

\begin{document}
\maketitle
\begin{abstract}
Current evaluations of large language models (LLMs) rely heavily on a growing collection of benchmarks and on aggregate benchmark scores, yet it remains unclear what this comparison actually captures, and what these scores reveal about models' underlying capabilities. 
Here, we propose a new paradigm for LLM evaluation, by asking whether benchmark performance reflects many independent abilities, or rather relies on a small number of shared dimensions. To answer this, we apply Factor Analysis (FA) to a massive performance  matrix of LLMs versus benchmarks \((60\times44)\)
revealing an \emph{intrinsically low-rank} structure of that matrix.
That is, a small number of latent factors captures most of the structure in the full task space.
This low-rank geometry reveals substantial redundancy across existing tasks and explains why many benchmarks appear to be measuring overlapping abilities.
We further show that these latent factors correspond to coherent, skill-like, dimensions of LLM behavior. 
Leveraging this latent skill-space, we deliver three practical tools for LLM evaluation and downstream users: (i)~identifying redundant tasks, (ii)~profiling new models using a small subset of tasks, and (iii)~selecting models aligned with desired skill profiles. Our method provides a solid alternative to the de-facto standard of a single aggregate score, and establishes an interpretable and practical framework for understanding and benchmarking LLM core capabilities.
\end{abstract}

\iftaclpubformat
\else
\fi

\section{Introduction} \label{sec:introduction}

Large Language Models (LLMs) now achieve state-of-the-art (SOTA) performance across a rapidly growing set of benchmarks, spanning question answering \cite{rajpurkar2016squad100000questionsmachine}, complex reasoning \cite{lyu2023faithfulchainofthoughtreasoning},  summarization \cite{hermann2015teachingmachinesreadcomprehend}, and many others. 

As these benchmarks proliferate, evaluation has increasingly relied 
on long lists of task-specific scores or on simple aggregate averages (e.g., Chatbot Arena by \citet{huggingface_lmsys_leaderboard}). 
Yet it remains unclear what such evaluations 
reveal about models’ underlying 
capabilities. 
Although individual datasets are designed to assess specific abilities, 
in practice tasks often combine multiple underlying skills, making it unclear how dataset scores relate to shared model behaviors.
This raises a set of fundamental questions: What latent information is captured by the diverse evaluation benchmarks? Moreover, to what extent do different datasets overlap in what they measure? And how can we meaningfully compare models for a given downstream case beyond aggregate scores?

In this paper, we propose a new, data-driven paradigm for LLM evaluation, inspired by psychometric theory. Specifically, we propose to treat benchmark tasks as test items, and LLMs as subjects, and to apply exploratory Factor Analysis (FA; \citet{thurstone1931multiple}) (\S\ref{sec:fa_method}) to empirical models' performance data. 
This analysis is designed to reveal whether LLM evaluation records exhibit a low-rank structure, in which a small number of latent factors can explain most of the variation in model performance across tasks.
This kind of structure would indicate that performance across many datasets is governed by a small number of shared dimensions, providing a compact view of model behavior.

To uncover this structure, our methodology proceeds in 
three stages:  
{\em comprehensive leaderboard construction}, {\em factor analysis}, and {\em skill naming}.  
Concretely, we first assemble a comprehensive leaderboard evaluating 60 diverse LLMs on 44 commonly used benchmarks, forming a new and diverse model–task performance matrix that serves as the basis for our analysis (\S\ref{sec:benchmarkLeaderboard}).
Then, we apply exploratory FA to this matrix to recover a small set of latent dimensions that summarize shared performance patterns across tasks. 
Lastly, we interpret each latent factor by analyzing its highest-loading tasks and assign it a descriptive label, yielding a set of coherent dimensions of model behavior. Throughout this paper, we refer to these labeled dimensions as {\em skills} (\S\ref{sec:results}).

\begin{figure}[t]
\centering
\includegraphics[width=0.9\linewidth]{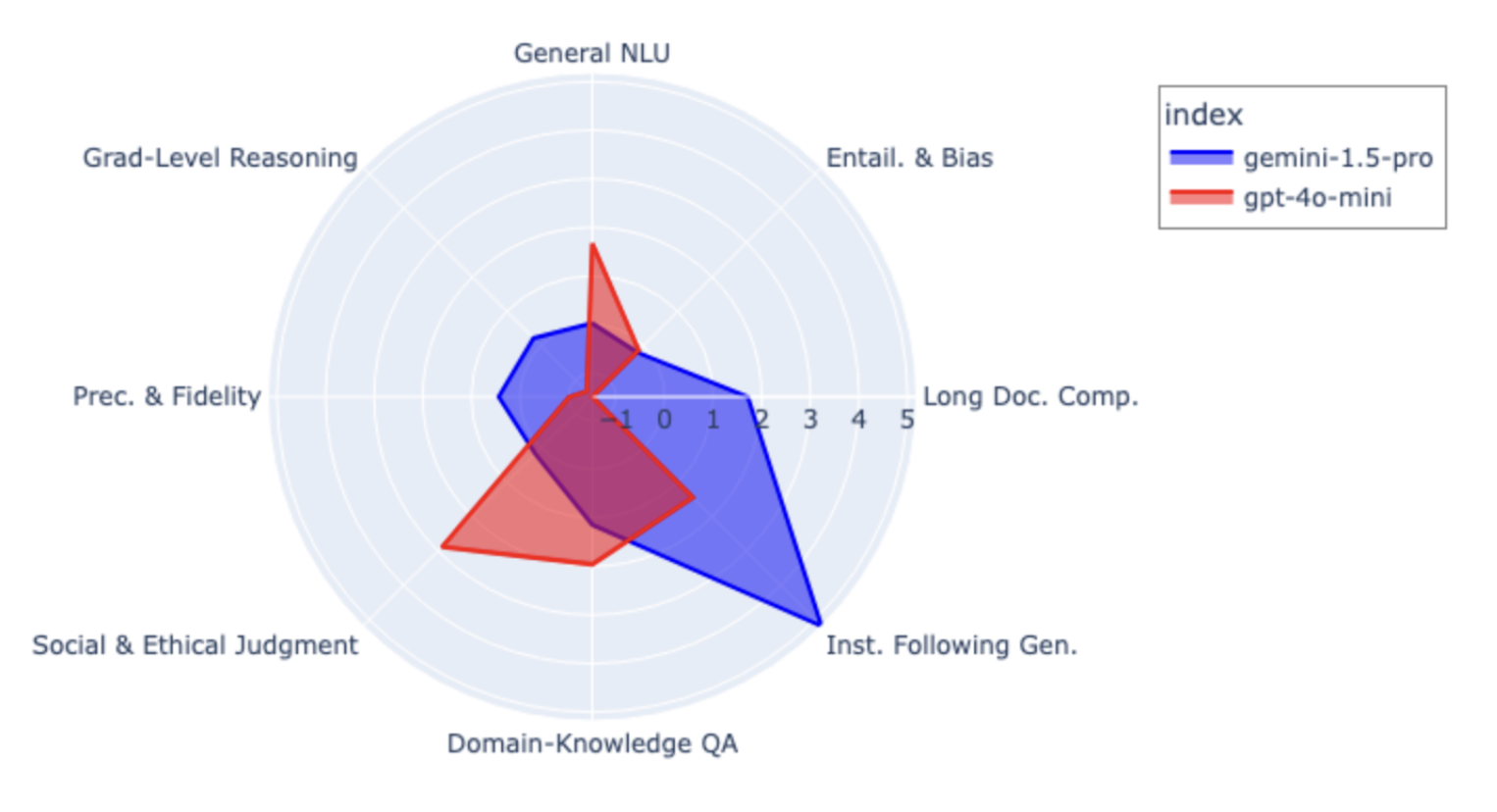}
\caption{
Despite similar Chatbot Arena~\cite{huggingface_lmsys_leaderboard} ratings (1322 vs.\ 1316), the models differ substantially in latent skill profiles, highlighting limitations of aggregate preference scores.
}
\label{fig:gpt-vs-gemini}
\end{figure}

The latent skill space reveals systematic differences across models, which are 
obscured by aggregate evaluation scores. 
For example, it reveals that models with similar Chatbot Arena rankings can occupy markedly different positions in the latent skill space, reflecting distinct strengths across the underlying dimensions (Figure~\ref{fig:gpt-vs-gemini}; \S\ref{subsec:arena}). 
We  show that the recovered latent structure is robust to 
task and model perturbations (\S\ref{subsec:validRobust}).
We also show that it generalizes to unseen datasets (\S\ref{subsec:newDatasets}).

Crucially, beyond the theoretical implications and the interpretation of models' scores, the existence of a low-rank latent skill space enables practical uses.
Subsequently, we deliver three automatic tools for practical applications: (i)~quantifying the novelty or redundancy of a new benchmark (\S\ref{subsec:newDatasets}), (ii)~efficiently profiling a new model from a small subset of tasks (\S\ref{subsec:NewModel}), and (iii)~selecting a model for unseen tasks by estimating their latent skill requirements (\S\ref{subsec:chooseModel}).
Together, these applications define a skill-centric approach to LLM evaluation that moves beyond opaque benchmark averages.

Taken together, our results establish FA as a principled and interpretable framework for understanding the latent structure of LLM evaluation data, giving rise to practical tools for more effective, efficient and far more transparent model assessment. We release our code, leaderboard, analytical matrices, and tools to facilitate further research on LLM evaluation and benchmarking.

\section{Proposal: Modeling Latent Skills}\label{sec:fa_method}

In this section we introduce our latent-skill modeling framework. We first motivate our approach, then we formalize our proposed factor-analytic approach for extracting shared performance dimensions from benchmark data, and lastly we describe how these dimensions may be interpreted.

\subsection{Methodological Motivation}

Our approach draws inspiration from classical psychometrics, which studies how latent constructs can be inferred from patterns of responses across many items. 
FA was developed precisely to separate \emph{shared} variance (interpreted as a common underlying ability) from \emph{item-specific} variance \cite{spearman1904}, and later became the foundation of widely used measurement frameworks, e.g., the Big Five personality traits \cite{Tupes1961, Costa1992}.

In our setting, tasks play the role of  {\em specific items} and LLMs  the role of {\em respondents}. The analogy is mathematical (rather than cognitive): 
FA provides a principled tool for identifying shared structure in model performance across tasks, by modeling cross-task covariance.
\enspace
Viewed through this lens, FA enables the isolation of latent dimensions that explain systematic performance variation across benchmarks, separated from task-specific noise.

\subsection{Formalization}
Our formal approach, in a nutshell, is as follows. We begin with a \textit{performance matrix} $\PerformanceMat$ (\S\ref{subsec:perfMatrix}): a $\textit{models}\times\textit{tasks}$ matrix that contains the scores of all tasks on all LLMs.
We then use \textit{Principal Axis Factoring} (PAF; \citealp{harman1976modern}), treating each task (column) as a variable and each model (row) as an observation. PAF extracts factors accounting for \emph{shared} variance across tasks while discarding task-specific noise (\S\ref{subsec:paf}). 
PAF yields three benefits:\footnote{Alternative decompositions such as PCA \cite{bishop2006prml} were considered; PCA is the closest baseline for variance-based dimensionality reduction.
However, while PCA also yields low‑dimensional representations, it maximizes the {\em total} variance, mixing {\em shared} and {\em unique} variation --- so a single high‑variance idiosyncratic feature can dominate. PAF instead models only shared variance across variables, making it more appropriate here. Our comparison in \S\ref{subsec:compPCA} confirms that PAF yields more stable and interpretable components.}
(i)~the resulting factors \textbf{generalize} more reliably to unseen tasks, since task-specific variance is removed;
(ii)~the solution is more \textbf{robust to outliers}, as variance unique to one task is treated as noise and cannot distort the factors; and
(iii)~the extracted factors are more \textbf{interpretable}, reflecting broad abilities rather than fragmented patterns. 
Finally, we present how we name the extracted latent factors as skills, and conceptualize an intuitive, interpretable {\em skill-based leaderboard} (\S\ref{subsec:comp2capab}).

\subsection{The Performance Matrix} \label{subsec:perfMatrix}

We first define a leaderboard-style performance matrix $\PerformanceMat$ to capture the systematic comparison of $M$ LLMs across $B$ diverse tasks. 
Constructing $\PerformanceMat$ involves (i) aggregating publicly available tasks, (ii) harmonizing heterogeneous evaluation metrics onto a unified 0–10 scale, and (iii) recording model–task scores. Further details on the actual construction of our leaderboard are provided in \S\ref{sec:benchmarkLeaderboard}.

\subsection{Principal Axis Factoring}\label{subsec:paf}

\paragraph{Generative view.} 
Given $M\times B$ performance matrix $\PerformanceMat$, PAF assumes factor decomposition model 
\begin{equation}
  \PerformanceMat \;=\; \ThetaMat\LambdaMat^{\!\top} + \epsilonMat,
  \label{eq:paf-matrix}
\end{equation}
with \(\ThetaMat \in\ \mathbb R^{M\times C}\) (\emph{skills} matrix),  
\(\LambdaMat \in \mathbb R^{B\times C}\) (task \emph{loadings}), and  
\(\epsilonMat \in \mathbb R^{M\times B}\) (task‑specific noise).

The {\em skills} matrix scores each model on each latent skill, and the
{\em loadings} are coefficients measuring how strongly a latent skill \(c\) contributes to the explanation of the performance of task \(b\).

\paragraph{Assumptions.}
For every model \(i\!\in\!M\), let \(\mathbf p_i \in \mathbb R^{B}\) denote its observed task‑performance vector:
\[
  \mathbf p_i := \bm{\theta}_i^{\top}\LambdaMat^{\!\top} + \bm{\varepsilon}_i^{\top} \in \mathbb R^{B},
\]
where \(\bm{\theta}_i^{\top}\) and \(\bm{\varepsilon}_i^{\top}\) are the \(i\)-th rows of latent skills and task‑specific noise. We treat \(\bm{\theta}_i\) and \(\bm{\varepsilon}_i\) as random vectors with:
\begin{align*}
  \mathbb E[\bm{\theta}_i] &= \mathbf 0, 
  &\operatorname{Cov}[\bm{\theta}_i] &= \boldsymbol{\Gamma} = \mathbf I_C \in \mathbb R^{C\times C}, \\
  \mathbb E[\bm{\varepsilon}_i] &= \mathbf 0, 
  &\operatorname{Cov}[\bm{\varepsilon}_i] &= \boldsymbol{\Psi}=\operatorname{diag}(\psi_1,\dots,\psi_B), \\
  & & \bm{\theta}_i &\perp \bm{\varepsilon}_i .
\end{align*}
We adopt the \emph{orthogonal‑factor} convention \(\boldsymbol{\Gamma}=\mathbf I_C\), so, each latent skill has unit variance, and skills are a-priori uncorrelated.

\paragraph{Population covariance.}
These assumptions give the factor‑analysis decomposition: 
\[
\mathbf R := \operatorname{Cov}[\bm p] = \LambdaMat\boldsymbol{\Gamma}\LambdaMat^{\!\top} +  \boldsymbol{\Psi} = \LambdaMat\LambdaMat^\top+\boldsymbol{\Psi},
\]
so that each off‑diagonal entry of \(\mathbf R\) is explained by the low‑rank term \(\LambdaMat\LambdaMat^{\!\top}\).  

The diagonal element \(\bigl(\LambdaMat\LambdaMat^{\!\top}\bigr)_{jj}\) is known as the \emph{communality} of task~\(j\):
\vspace{-2pt}
\[
  h_j^{2}\;=\;\sum_{c=1}^{C}\lambda_{jc}^{2},
\]
\vspace{-2pt}
namely the portion of its variance explained by the common factors.  
The remainder \(\psi_j=1-h_j^{2}\) is its \emph{uniqueness} (task‑specific variance). 

The loading matrix \(\boldsymbol{\Lambda}\) is identifiable only up to rotation, so PAF focuses on the \emph{subspace} it spans. Orthogonal rotations such as Varimax maximize sparsity and improve interpretability.

\paragraph{Learning objective.}
Given \(\mathbf R\) and a chosen number of skills \(C\ll B\), PAF
estimates \((\widehat\LambdaMat,\widehat{\boldsymbol{\Psi}})\) by
\begin{equation}
  \min_{\LambdaMat,\;\boldsymbol{\Psi}\text{ diagonal}}
  \bigl\|\mathbf R-\LambdaMat\LambdaMat^{\!\top}-\boldsymbol{\Psi}\bigr\|_F^{2}.
\end{equation}
See \cref{app:paf} for the iterative algorithm used to estimate \(\LambdaMat\) and \(\boldsymbol{\Psi}\). 

\paragraph{Representing a new model.}

Given a new model with task scores \(\mathbf p_{\text{new}}\!\in\!\mathbb R^{B}\)\footnote{$z$‑scored using the task means and standard deviations of the original performance matrix.}, we compute its skill vector using \emph{regression (Thomson) scores}, which minimize \(\operatorname{MSE}=\mathbb E[\|\widehat{\bm\theta}-\bm\theta\|_2^{2}]\). 
This involves two steps: regress the latent skills from FA (\(\ThetaMat\)) on the observed scores (\(\PerformanceMat\)) via population multiple-OLS and then apply \(\mathbf B_{\text{reg}}\) to the new task vector:
\begin{equation}\label{eq:thomson-BLP}
  \mathbf B_{\text{reg}}^{\!\top} = \LambdaMat^{\!\top}\mathbf R^{-1},
  \qquad
  \widehat{\bm\theta}_{\text{reg}} = \mathbf B_{\text{reg}}^{\!\top}\,\mathbf p_{\text{new}}.
\end{equation}

Note that because 
\(\mathbf B_{\text{reg}}\) depends only on \(\widehat\LambdaMat\) and \(\widehat{\boldsymbol{\Psi}}\), it can be stored once and reused for any new model via a single matrix–vector product.

\subsection{Labeling Factors as Skills}\label{subsec:comp2capab}

PAF yields $C$ latent factors capturing shared structure across tasks. To make them interpretable, we assign each factor a concise label based on its highest-loading tasks.
Note that factor labels are used only for exposition; all analyses depend 
on the learned {\em loadings} and scores.

Factor interpretation is grounded in the semantic commonalities among the highest-loading tasks for each factor, which serve as diagnostic indicators of the underlying dimension. To support scalability and to reduce individual subjective biases, we employ an LLM-assisted labeling procedure that proposes names for each factor given short summaries of its representative tasks. 
The procedure is agnostic to any pre-defined skill taxonomy.

Factor labels are interpretive summaries of the loading structure and do not affect the factors.
\footnote{
We verified empirically (appendix~\ref{app:skillNaming}) that labels are consistent across judge models and prompts. Prompt sensitivity remains a limitation, which we leave for future work.}

We present the named latent dimensions in \S\ref{subsec:numskills} and refer to them as \emph{skills} throughout the paper.
The role of labeling is interpretive: it provides a concise semantic summary of each latent dimension uncovered by FA, but does not influence factor extraction, loadings, or model scores. All quantitative results and robustness analyses are therefore independent of the specific phrasing used to describe each factor.
Full details of the naming procedure are provided in \cref{app:skillNaming}. 

\section{The Task-Based Leaderboard} \label{sec:benchmarkLeaderboard}

Developers face major challenges in reconciling diverse evaluation protocols, including diverse metrics and task formats, and in ensuring scalability and reproducibility. 
To mitigate this, we deliver a standardized, transparent leaderboard comparing LLMs across {\em skills}, rather than tasks, highlighting their current strengths and limitations, and guiding future research. 
Concretely, we compile a matrix of 60 models evaluated on 44 tasks, which forms the empirical basis for our FA.
Here, we describe the construction of the performance matrix $\PerformanceMat$.

\paragraph{Tasks.}\label{sec:dataset}

We evaluate models on a suite of $B=44$ publicly available tasks, balancing classification and generation tasks across short- and long-context settings.
Of these, 25 are classification-based: 13 binary (e.g., QNLI, BoolQ) and 12 multiple-choice (e.g., MMLU, MedQA). 
The remaining 19 generation tasks cover tasks such as summarization (XSum), open-domain QA (TriviaQA), and dialogue (DailyDialog), and span different domains such as legal, medical, and conversational. The complete list of tasks is provided in ~\cref{App:fullDataset}.
Integrating these diverse tasks into a single leaderboard enables comprehensive model evaluation via the method described in \S\ref{subsec:paf}.

\paragraph{Prompts.}

For fair and robust evaluation across diverse models, we designed three prompts per task that produced valid outputs across models (see~\cref{App:BenchmarkInstruction}). Task scores average performance over all three prompt variants per task. 

Averaging across prompts reduces variance but may attenuate covariance for prompt-sensitive tasks, slightly weakening correlations. In practice, the dominant factor structure remains stable, suggesting limited impact on the learned skill space.

\paragraph{Metrics.}
For all classification tasks we used standard metrics, prioritizing the most common metric when multiple exist. For generation tasks, we adopted an LLM-as-a-judge approach \cite{zheng2023judgingllmasajudgemtbenchchatbot} using Deepseek-R1 \citep{deepseekai2025deepseekr1incentivizingreasoningcapability}. Full detailing of the tasks and evaluation method are provided in~\cref{App:fullDataset}.

\paragraph{Normalization and Standardization.}
All task scores were normalized to [0, 1] and then standardized (zero mean, unit variance) before FA.  
Because FA uses correlations, results are invariant to these linear transformations.

\paragraph{Models.} 

We compile a leaderboard spanning all tasks on $M=60$ instruction-tuned LLMs from 24 model families, including both open-source (e.g., LLaMA) and proprietary systems (e.g., Gemini).
Most models (53/60) use decoder-only architectures, while the remainder follow an encoder–decoder design. Open-source parameter counts range from 2B to 27B; proprietary sizes are largely undisclosed. The set includes models from commercial labs (e.g., GPT), academia (e.g., Bloom), and startups (e.g., DeepSeek), with several RLHF variants (e.g., Claude).

\section{Results: A Skill-Based Leaderboard}\label{sec:results}

\subsection{The Latent Skills}\label{subsec:numskills}
We applied PAF to the model–task performance matrix we constructed in   \textsection\ref{sec:benchmarkLeaderboard}, and uncovered 8 latent dimensions, labeled using the procedure in \S\ref{subsec:comp2capab}. We refer to these dimensions as {\em skills}.
The recovered skills are: {\em (1) General NLU, (2) Entailment \& bias, (3) Long‑document comprehension, (4) Instruction following \& generation, (5) Domain QA, (6) Social \& ethical judgment, (7) Precision \& fidelity} (i.e., exact adherence to reference outputs, numerical values or factual details), and {\em (8)~Grad‑level reasoning}. Table~\ref{tab:factor_loadings} lists each factor with its name, description, and most strongly associated tasks.

Assigning human‑interpretable names to each skill relied on a two‑step procedure (\S\ref{subsec:comp2capab}) which leverages GPT‑4o and DeepSeek‑R1~\citep{deepseekai2025deepseekr1incentivizingreasoningcapability} to summarize the most influential tasks. The number of latent factors ($C$) was chosen via Kaiser’s $\lambda>1$ rule and an 85\% cumulative variance threshold (Fig.\ref{Fig:fa_n_factors_explained} in \cref{subapp:numFactors})~\citep{kaiser1960application,harman1976modern}, ensuring that each dimension retained shared variance without overfitting noise.

Note that the resulting granularity reflects a balance between statistical support, robustness, and interpretability for the current task suite --- rather than a claim of a unique ``correct'' skill resolution.
Also note that we use the term \emph{skill} to descriptively refer to a latent dimension of model behavior, rather to refer to, e.g., a pre-defined psychological trait.

\begin{figure*}[t]
\centering
\includegraphics[width=1\textwidth]{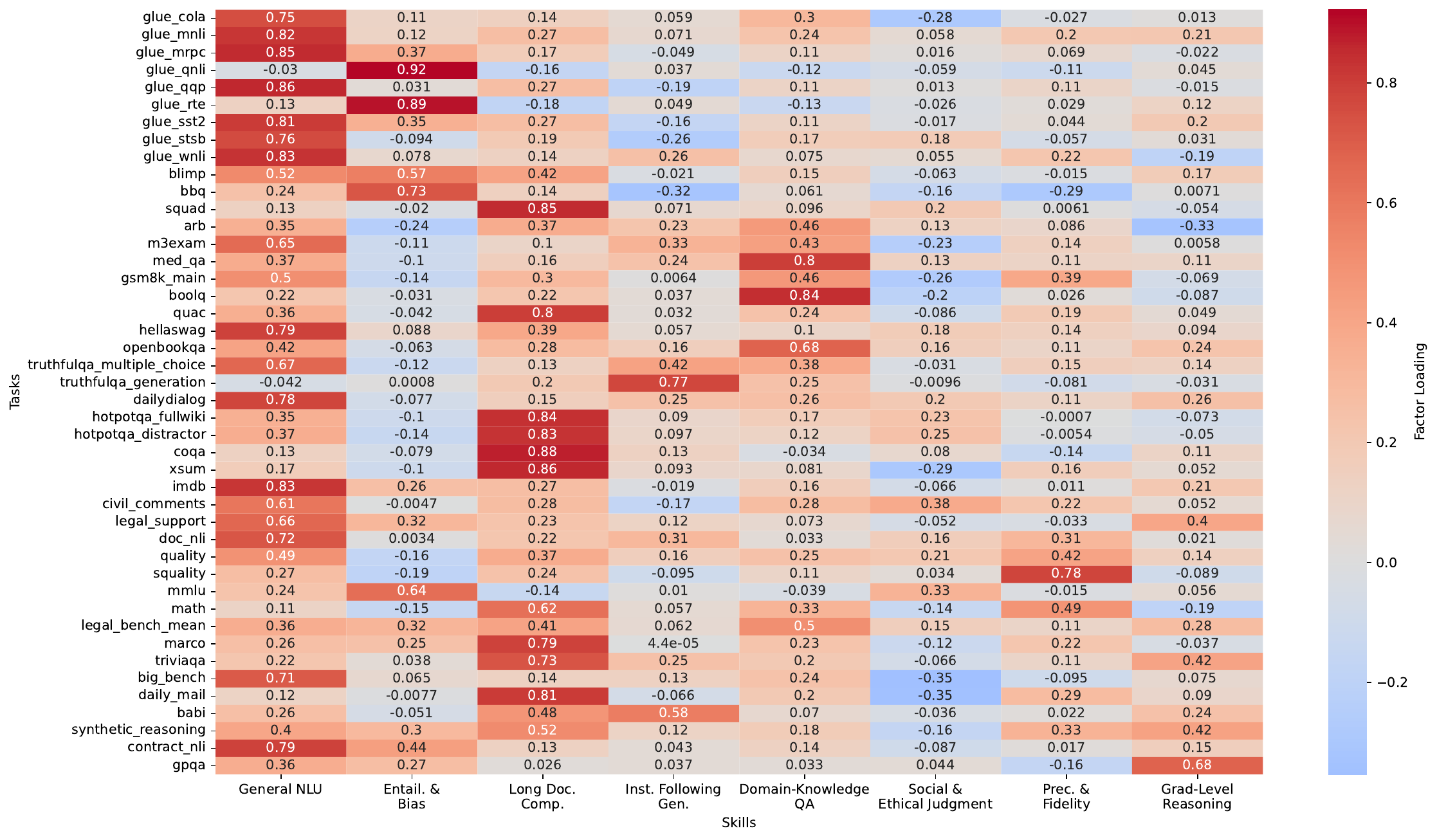}
\caption{Task–factor loading matrix ($\LambdaMat$) for 8 skills. Larger value indicate stronger association; positive values mean the task reflects the skill, negative values suggest an inverse pattern.}
\label{Fig:factor_loading_8_factors}
\end{figure*}

\begin{table*}[t]
\centering
\small            
\setlength{\tabcolsep}{4pt}
\resizebox{0.95\textwidth}{!}{
\begin{tabular}{@{}p{4cm}p{3cm}p{4.8cm}p{6.3cm}p{3.2cm}@{}}
\toprule
\textbf{Factor Name} & \textbf{Factor Nickname} & \textbf{Description} & \textbf{Strong $+$ tasks (loading)} & \textbf{Strong $-$ tasks (loading)} \\
\midrule
1.\; General NLU & \FactorName{General NLU} &
Everyday syntax, semantics, sentiment classification. &
QQP (.86), MRPC (.85), WNLI (.83), IMDB (.83), MNLI (.82), SST-2 (.81), HellaSwag (.79), DailyDialog (.78), Contract-NLI (.78) &
\\

2.\;Fine-Grained Entailment\ \& Token Bias & \FactorName{Entailment \& Bias} &
Short entailment and token-level bias/grammar probes. &
QNLI (.92), RTE (.89), BBQ (.73), MMLU (.64), BLiMP (.57) &
\\

3.\; Document Reading, Retrieval \& Summarization & \FactorName{Long Doc. Comprehension} & 
Long-context reading, retrieval, and abstractive summarization. &
CoQA (.88), XSum (.86), SQuAD (.85), Hotpot-FW (.84), Hotpot-Distr. (.83), CNN/DM (.81), QuAC (.80), MS MARCO (.79) &
\\

4.\; Truthful Instruction\ Generation &\FactorName{Inst. Following Gen.} & 
Truthful, instruction-following open-ended generation. &
TruthfulQA-Gen (.77), bAbI (.58), M3Exam (.33) &
BBQ (\textminus.32)
\\

5.\; Specialized Domain-Knowledge QA & \FactorName{Domain QA} &
Expert factual QA in medical, legal, scientific domains. &
BoolQ (.84), MedQA (.80), OpenBookQA (.68), LegalBench (.50), GSM8K (.46) &
\\

6.\; Social / Ethical Judgment &\FactorName{Social / Ethical Judgment} & Social and ethical judgment vs.\ news-style summarization. &
CivilComments (.38), MMLU (.33) &
BigBench (\textminus.35), DailyMail (\textminus.34) 
\\

7.\; Faithful Summarization \& Quantitative Precision & \FactorName{Prec. \& Fidelity} & Token-level fidelity and quantitative precision. &
SQuALity (.78), Math (.49), Quality(.41),  GSM8K (.39), Synth-Reason (.33) &
BBQ (\textminus.29) 
\\

8.\;Grad-Level Scientific and Legal Reasoning & \FactorName{Grad-Level Reasoning} & 
Graduate-level scientific and legal reasoning. &
GPQA (.68), Synth-Reason (.42), TriviaQA (.42), Legal-Support (.40) &
ARB (\textminus.33) 
\\ 
\bottomrule
\end{tabular}}  
\caption{Latent skill factors, with names, short descriptors, and the most strongly associated task loadings.
}
\label{tab:factor_loadings}
\end{table*}

\subsection{The Skill-Based Leaderboard}\label{subsec:skillsleaderboard}

\begin{figure*}[!t]
\centering
\includegraphics[width=\linewidth]{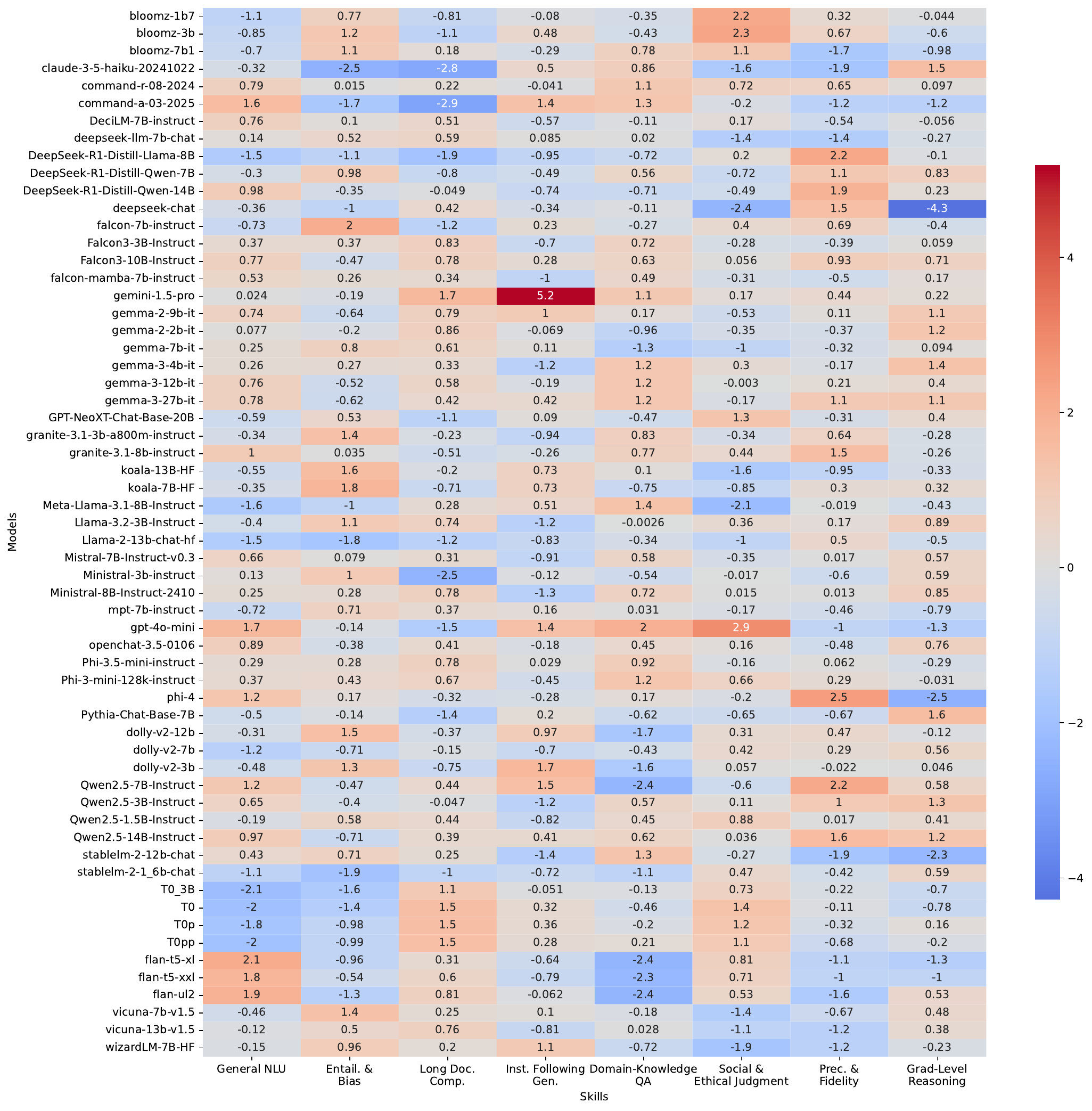}
\caption{Model scores on 8 latent skills. Higher values indicate stronger skill performance; positive = above average, negative = below average.}
\label{fig:factor_scores_ours}
\end{figure*}

To evaluate models along these dimensions, we construct a skill-based leaderboard from the score matrix~$\ThetaMat$ (model–skill relationships; Fig.~\ref{fig:factor_scores_ours}). Unlike traditional task-level leaderboards, which report average performance across benchmarks, this representation ranks models by skill profiles, offering a far more compact, structured view of abilities.

\subsection{The Analysis of Latent Skills}

We analyze the 
uncovered structure by examining how tasks align or separate in the learned skill space. In particular, we focus on cases where tasks that appear similar by format or domain diverge across factors, as well as cases where superficially different tasks converge on the same  dimension.

\paragraph{Surface-similar tasks diverge.}
First, let us consider pairs of tasks that share a format yet fall on different factors. For example, GPQA and TriviaQA are both defined as question-answering tasks, but GPQA loads on Factor 8.\ \FactorName{Grad.-Level Reasoning}, because it stresses physics derivations, whereas TriviaQA lands on Factor 3.\ \FactorName{Long Doc.\ Comp.}, wherein evidence extraction dominates; it appears that the \emph{skills} needed for question-answering differ between these datasets. A similar split appears inside GLUE: MRPC (paraphrase detection) sits with Factor 1’s \FactorName{General NLU}, but its sibling QNLI (question–sentence entailment) migrates to Factor 2, reflecting the finer-grained inference needed for entailment. 

Generative tasks diverge as well: CoQA (conversational, document-grounded QA) joins Factor 3 \FactorName{Long Doc.\ Comp.}, while TruthfulQA-Generation, whose main challenge is hallucination avoidance, aligns with Factor 4’s \FactorName{Inst.-Following Gen.} Even two entailment datasets separate --- GLUE’s MNLI stays in Factor 1, but QNLI heads to Factor 2 --- highlighting how increased reasoning granularity reshapes the latent map.

\paragraph{Surface-dissimilar tasks converge.}
The reverse pattern also appears: tasks with different formats or domains can load on the same factor due to shared underlying demands. For example, SQuALity (faithful-summary scoring) and Math (symbolic arithmetic) both rely on token-level exactness; unsurprisingly, they co-load on Factor 7, our \FactorName{Prec. \& Fidelity} axis. 
SQuAD and XSum, 
extractive versus abstractive summarization, meet in Factor 3 because both demand document comprehension and content selection.
Finally, news-domain MRPC and legal Contract-NLI share Factor 1, \FactorName{General NLU}, suggesting that their relative simplicity (and high model scores) overrides domain differences.

Taken together, these examples illustrate that the latent factors organize tasks according to shared behavioral demands rather than surface characteristics such as format or domain.

\paragraph{Breadth and interpretability of the latent skills.}

As summarized in Table~\ref{tab:factor_loadings}, the eight recovered skills differ in both breadth and diagnostic coverage across benchmarks. 
Some, such as \FactorName{General NLU} and \FactorName{Domain QA}, integrate a broad family of tasks, whereas others --- e.g., \FactorName{Prec. \& Fidelity} and \FactorName{Ethical Judgment} --- are defined by a smaller set of highly discriminative tasks that capture specialized competencies. 
Negative loadings are expected within such latent dimensions: they indicate inverse co-variation, where excelling on one group of tasks (e.g., factual precision) often accompanies lower performance on another emphasizing complementary demands (e.g., stylistic alignment or social sensitivity). 
Hence, the sign of a loading indicates relative emphasis along a shared latent dimension, not whether a model is better or worse overall.

\section{Evaluation of the Uncovered Skill Set}\label{sec:faValidity}

In this section, we aim to evaluate the validity and interpretability of the uncovered skill dimensions. To this end, we address two key questions: 

\begin{itemize}
    \item 
    [
    (i)]
    ~Does the factor structure reflect a stable and internally coherent latent space?
    
    \item 
    [
    (ii)]
    ~Do the derived skill scores align with external (human) judgments of model behavior?
\end{itemize}

To answer (i), \S\ref{subsec:validRobust} draws on standard FA metrics \cite{harman1976modern, CronbachMeehl1955}, measuring internal consistency, 
stability under task perturbation, 
and factor specificity (uniqueness, outlier detection). 
We also compare our results using the PAF method to results using PCA for dimension redundancy.
To address (ii), \S\ref{subsec:arena} compares our skill-based rankings to pairwise human preferences from Chatbot Arena \cite{huggingface_lmsys_leaderboard}, testing whether aggregated human judgments align with models’ inferred skills.

\subsection{Latent Space Validation and Reliability}\label{subsec:validRobust}

We evaluated the inherent quality of the identified skills by assessing the latent skill space's coverage (uniqueness) and internal consistency (reliability).

\paragraph{Uniqueness.}
Uniqueness (\(u_j^2 = 1 - h_j^2\)) quantifies how much of a task’s variance is \emph{not} captured by latent factors. Low values show the skill space adequately explains task behavior, i.e., tasks are well-covered by the factor model.
All tasks in our eight-factor solution (Fig.~\ref{Fig:uniqueness} in\cref{subapp:uniqueness}) sit below the $\sim$0.40 threshold (max: \textsc{MMLU}\,0.38), indicating strong coverage~\citep{Costello2005}.

\paragraph{Reliability.}

To confirm that tasks grouped under each factor measure the same underlying skill, we assessed their \emph{internal consistency} --- a standard proxy for {reliability}. For each factor, we selected tasks with relatively high loadings (z-score $>\!0.8$; minimum four tasks).
Using the resulting model-score matrix (models × tasks), we computed inter-task correlations and evaluated consistency via Cronbach’s~$\alpha$ \cite{Cronbach1951} and McDonald’s~$\omega$ \cite{McDonald1999}.\footnote{Cronbach’s~$\alpha$ assumes equal contributions (tau-equivalence) of all items to the latent factor, while McDonald’s~$\omega$ accounts for differing loadings and is generally more reliable in factor models~\cite{McDonald1999}.}
Both statistics exceeded the conventional 0.80 threshold ($\alpha=0.836$, $\omega=0.821$), showing models behave consistently across tasks within each factor --- evidence that the factors are \textbf{reliable} measures of latent skills.

\paragraph{Robustness of the latent-skill space.}

In this section we verify that the eight‐factor solution is not a fragile artifact of any single modeling choice.  
To this end, we apply three stress tests: (i) \textbf{factor dimensionality} --- refitting the model at \(k\!=\!6,7\) (under-extraction) and \(k\!=\!9,10\) (over-extraction); (ii) \textbf{model subsampling} --- five runs in which 30 \% of the LMs were held out at random; (iii) \textbf{task removal} --- leave-one-task-out in the PAF fit.

We aligned each perturbed loading matrix to the reference solution using orthogonal procrustes analysis~\cite{Schönemann66}, and deviation was measured with principal angles.\footnote{Principal angles quantify the minimal angular deviation between corresponding subspaces (\(A\) and \(B\)) of two factor solutions, defined by $\cos(\psi) = \max_{u \in A, v \in B} \frac{u^\top v}{\|u\|\|v\|}$. Values $\psi \le 25^\circ$ suggest strong factor similarity~\citep{Tucker1951}.} Tucker’s congruence coefficient (\(\varphi\)) provided a complementary similarity check.\footnote{Tucker’s congruence coefficient \(\varphi\) measures the cosine similarity between two factor-loading vectors \(\ell_a\) and \(\ell_b\): $\varphi = \frac{\sum_j \ell_{a,j} \ell_{b,j}}{\sqrt{\sum_j \ell_{a,j}^2} \sqrt{\sum_j \ell_{b,j}^2}}$.} We expect high congruence coefficient and low principal angles if the factor space after each ablation is similar to the original one. 

Across perturbation analyses, two complementary findings emerge: varying the number of factors confirms that the eight-factor model captures the data without over- or under-fitting, while removing data (models or tasks) shows that this space is remarkably stable. 
Specifically, under-extraction (\(k=6,7\)) caused skill merging or loss (principal angles \(43^\circ\!-\!60^\circ\)), whereas over-extraction (\(k=9,10\)) produced redundant, near-orthogonal axes while the eight core factors rotated by at most \(26^\circ\). 
When 30\% of models were removed, every factor re-emerged with low alignment error (\(<14^{\circ}\), s.d.\!\(<8^{\circ}\)), and leave-one-task-out runs yielded minimal rotation (\(\le 2.4^\circ\)). 
Together, these results confirm that the eight-factor solution 
captures a set of distinct, stable capabilities and remains robust to realistic data perturbations.

Together, these results confirm that the eight-factor solution captures a set of distinct, stable capabilities and remains robust to data perturbations.

\paragraph{Models outlier detection.}\label{subsec:MahalanobisTest}
To verify that all current models lie within the latent-skill space, we also computed Mahalanobis distances\footnote{an `elliptical' metric accounting for feature covariance, is defined as $d_{M}(\boldsymbol\phi) =\sqrt{(\boldsymbol\phi - \bar{\boldsymbol\phi})^\top \Sigma^{-1} (\boldsymbol\phi - \bar{\boldsymbol\phi})}$, where \(\bar{\boldsymbol\phi}\) and \(\Sigma\) are the sample mean and covariance.}
~\citep{mahalanobis1936generalised} from the distribution of other models; 
none exceeded the 99.5th percentile of the \(\chi^2_B\) distribution~\citep{de_maesschalck2000mahalanobis}, indicating no outliers.
Overall, the 
skill space is structurally stable under design choices, without statistical outliers.

\paragraph{PCA vs.\ PAF}\label{subsec:compPCA}

Principal Axis Factoring (PAF), the method we use, and Principal Component Analysis (PCA), a viable alternative, differ fundamentally in what they model: PAF explicitly isolates \emph{shared variance} across tasks, whereas PCA maximizes \emph{total variance} and does not distinguish shared structure from task-specific effects. 
To assess whether this distinction matters empirically, we repeated all robustness, interpretability, and downstream analyses using PCA. 
The results show that PCA yields a substantively different and less useful latent representation (Table~\ref{tab:paf-vs-pca-summary}).

First, PCA recovers a different latent representation than PAF. 
Beyond the strongest global direction, alignment between PCA components and PAF factors is limited (Tucker congruence $\varphi = 0.29\!-\!0.93$), and model rankings along corresponding dimensions differ substantially, as evidenced by only moderate correlation between PCA- and PAF-derived model scores (Spearman $r = 0.49\!-\!0.73$). 
Furthermore, this difference is reflected in external validation. 
As will be seen shortly in \S\ref{subsec:arena}, the aggregate PAF skill score aligns strongly with human preference judgments in Chatbot Arena ($\rho = 0.73, p < 0.005$), whereas the aggregate PCA score shows no reliable correlation ($\rho = 0.28, p > 0.3$).
We interpret this gap as a consequence of PCA’s objective: by maximizing total variance, PCA mixes shared performance structure with task-specific variation, which limits its ability to recover the latent capabilities that consistently influence human preference judgments.

Second, these differences reflect PCA’s objective rather than genuine recovery of shared task structure. 
Because PCA maximizes total variance, its components receive small contributions from many tasks, resulting in diffuse loadings.
Consequently, removing individual tasks or subsampling models has little effect on the solution.
This insensitivity reflects PCA’s focus on global variance directions, not the identification of latent dimensions supported by consistent task co-variation.
Third, PCA components are less interpretable and less useful for evaluation. 
Several PCA components are supported by a single dominant task, making standard reliability measures (Cronbach’s $\alpha$, McDonald’s $\omega$) undefined, whereas PAF factors consistently exhibit high internal consistency ($\alpha, \omega > 0.80$). 

Finally, when we apply the same downstream reconstruction and prediction pipelines 
we used for task novelty assessment, new-model profiling, and model selection (\S\ref{sec:tools}) to PCA components, rather than PAF factors, performance degrades substantially: reconstruction error increases and predictions for unseen models or tasks become unreliable.

Taken together, this comparison shows that while PCA identifies directions of maximal variance, it conflates shared and task-specific effects and yields latent dimensions that are less interpretable, reliable, and predictive.
PAF more directly recovers shared performance structure across tasks, justifying its use as a principled latent-skill model.

\begin{table*}[t]
\centering
\small
\begin{tabular}{p{4.4cm} p{5.2cm} p{5.2cm}}
\toprule
\textbf{Criterion} & \textbf{PAF (ours)} & \textbf{PCA} \\
\midrule
\textbf{Variance treatment}
& Models shared variance and estimates task-specific variance 
& Maximizes total variance without separating task-specific effects \\[0.4em]

\textbf{Alignment with PAF factor structure} 
& Reference latent structure 
& Limited alignment beyond first component (Tucker $\varphi=0.29$--$0.93$) \\[0.4em]

\textbf{Agreement in model rankings (PAF vs.\ PCA)} 
& Stable across re-fits 
& Moderate agreement with PAF ($r=0.49$--$0.73$) \\[0.4em]

\textbf{Loading concentration} 
& Sparse, multi-task loadings defining each factor 
& Broad loadings spread across many tasks \\[0.4em]

\textbf{Reliability ($\alpha,\omega$)} 
& High; multi-task latent skills 
& Often undefined (components dominated by a single task) \\[0.4em]

\textbf{Structural sensitivity (resampling \& dimensionality)} 
& Meaningful factor merging/splitting and interpretable changes 
& Minimal change due to variance maximization \\[0.4em]

\textbf{Interpretability}
& Shared latent capabilities 
& Maximal variance directions \\[0.4em]

\textbf{Correlation reconstruction} 
& Lower error 
& Higher error \\[0.4em]

\textbf{Arena alignment} 
& Strong ($\rho=0.73$, $p<0.005$) 
& None ($\rho=0.28$, $p>0.3$) \\
\bottomrule
\end{tabular}
\caption{Comparison of PAF and PCA across alignment, robustness, interpretability, and external validation criteria.}
\label{tab:paf-vs-pca-summary}
\end{table*}

\subsection{Comparison with Human Preferences}
\label{subsec:arena}

Chatbot Arena~\cite{huggingface_lmsys_leaderboard} provides large-scale human 
judgments and a single global Elo score~\cite{elo1978rating}\footnote{Adapted from competitive games, this dynamic rating system updates based on blind A/B test outcomes.} per model, and is often treated as a proxy for overall 
quality. 
We therefore use Arena as an external reference for validating our multidimensional skill scores.
We find that while aggregating out skill scores aligns strongly with Arena Elo, individual skills contribute unevenly to this alignment --- showing that Arena reflects only a subset of model skills. As a result, models with similar Elo scores can differ substantially in underlying strengths, motivating skill-aware evaluation beyond a single global preference.

\paragraph{Overall correlation.}

To test whether our skill space captures what Arena raters reward at a global level, we correlate Arena Elo ranking with the average of our eight latent skill scores for the 13 models shared across both settings. This yields a strong Spearman correlation (\(\rho = 0.73, p < 0.005\)). A leave-one-out analysis over 12-model subsets gives \(\rho = 0.722 \pm 0.058\), indicating that the agreement is stable under small changes in the model set.

\paragraph{Task-level baselines.}

To contextualize this alignment, we evaluate task-level baselines. 
A simple average of task scores correlates strongly with Arena Elo ($\rho \approx 0.86$), but this alignment is driven by a narrow tasks subset. 
The most Arena-aligned task, \texttt{HotpotQA-FullWiki}, achieves near-perfect correlation ($\rho \approx 0.97$), and averaging the top-5 Arena-correlated tasks yields a similarly strong baseline.

These results indicate that Arena preferences are heavily influenced by specific interaction styles—particularly multi-hop QA and short-form reasoning—rather than reflecting a balanced assessment of model capabilities. 
While task-level aggregation can reproduce Arena rankings, it relies disproportionately on a small subset of highly Arena-aligned tasks.

By contrast, the latent-skill framework groups tasks into stable dimensions that capture shared behavioral structure. 
This abstraction reveals both where human preferences align with underlying capabilities and where they diverge, yielding a more interpretable and generalizable evaluation.

\paragraph{Skill–level alignment.}

Although the aggregate skill score correlates with Arena Elo, individual skills exhibit systematically different relationships with human preference judgments (Table~\ref{tab:arena-skill-corr} in \cref{subapp:arenastd}). 
To characterize these differences, we analyze both each skill’s-Arena correlation, and the correlations of its highest-loading tasks. This reveals 4 
patterns.

(1) \emph{Arena-rewarded skills} (Case~A), such as \FactorName{General NLU} and \FactorName{Domain QA}, show strong positive correlations, and their highest-loading tasks are also Arena-aligned. This suggests that broad comprehension and confident factual recall are consistently salient to raters.

(2) \emph{Under-recognized skills} (Case~B), including \FactorName{Prec.\ \& Fidelity} and \FactorName{Grad-Level Reasoning}, exhibit weak skill-level correlation despite high internal consistency. Tasks loading on these skills emphasize numerical exactness, reference faithfulness, and multi-step reasoning, and show heterogeneous Arena alignment—some correlating strongly (e.g., GSM8K), others weakly (e.g., GPQA)—yet they co-vary reliably across models. As shown in \S\ref{subsec:validRobust}, these skills are internally coherent ($\alpha>0.85$, $\omega>0.80$). Their weak Arena alignment therefore reflects a limitation of preference-based evaluation, which favors fluency, brevity, and apparent confidence over capabilities that require careful verification, such as factual precision or deep reasoning.

(3) \emph{Context-dependent skills} (Case~C), such as \FactorName{Long-Document Comprehension}, \FactorName{Instruction Following / Generation}, and \FactorName{Social–Ethical Judgment}, are rewarded only when Arena prompts elicit the relevant capability. For example, \FactorName{Long-Document Comprehension} includes \texttt{HotpotQA-FullWiki}, which correlates strongly with Arena, but also aggregates long-context and summarization tasks that are rarely exercised in Arena-style interactions. Reduced skill-level alignment thus reflects limited prompt coverage in preference comparisons, rather than insensitivity of human raters to these capabilities.

(4) \emph{Arena-penalized} skills (Case~D), such as \FactorName{Entailment \& Bias}, emphasize formal entailment and bias avoidance, often yielding cautious or rigid responses that are often judged as less helpful in preference-based comparisons.

Taken together, these results show that Arena reflects only a subset of the latent skill space. Human preferences consistently reward general comprehension and confident recall, while underweighting or penalizing deep reasoning, factual precision, and bias mitigation. Skill-based evaluation therefore complements preference-based metrics by explaining which capabilities are visible, overlooked, or penalized under current paradigms.

\paragraph{Illustrative cases.}

To illustrate these effects, Fig.~\ref{fig:gpt-vs-gemini} compares two models with nearly identical Arena Elo scores. Despite similar 
rankings, the models differ substantially across 
latent skills, revealing complementary strengths 
invisible to aggregate evaluation. This example shows how skill-based analysis enables more informed model selection than reliance on a single global score.

\section{Applications: Skill-Based Evaluation}\label{sec:tools}

Beyond descriptive analysis, the latent skill space enables practical tools that address concrete evaluation challenges faced by LLM developers and users.
Rather than asking how a model performs on dozens of benchmarks, our framework supports targeted questions such as:
\emph{Is this new benchmark actually novel?},
\emph{How can I characterize a new model without running a full evaluation suite?}, and
\emph{Which existing model is best suited for a new task?}

We instantiate this perspective through a set of applications:
(i) diagnosing whether a new dataset adds genuine signal or largely re-measures existing skills;
(ii) profiling an unseen model from a small, diverse subset of tasks; and
(iii) selecting the best model for a new task with minimal evaluation cost.
Each application is validated on held-out data.
We further examine how the skill space evolves with new models and how it can be maintained under distribution shift.

We separate (i) methodological diagnostics (Sections~\ref{subsec:newDatasets}-~\ref{subsec:chooseModel}) from (ii) empirical validation (Section~\ref{subsec:maintaining}).
All tools rely on the same structural property: the model\(\times\)task performance matrix is well-approximated by a low-dimensional skill space.
This low-rank structure makes it possible to infer missing entries reliably.
Estimating a new model corresponds to \emph{row completion} (inferring a model’s skill profile from a few task scores). Analyzing a new task corresponds to \emph{column completion} (inferring task loadings from a few model evaluations).\footnote{All applications depend on the shared-variance structure recovered by PAF; analogous pipelines using PCA yield substantially higher error and instability (\S\ref{subsec:compPCA}).}

If new data falls outside the skill space, we re-estimate the factor model on the expanded matrix; this requires additional evaluations but is straightforward.
\footnote{Unless otherwise stated, numerical thresholds in this section are empirically chosen for illustration and validated in the experimental settings below.}

\subsection{Is this Task Redundant or Novel?}\label{subsec:newDatasets}

The growing number of evaluation datasets raises the question whether a new task adds genuinely new information or merely re-measures existing skills. To answer this,
we place each task on a \emph{novel--redundant} continuum using 3 diagnostics: 

 (1) Raw Score Redundancy: 
    We test if a task’s performance vector is predictable from others. High collinearity (max Pearson \(r > 0.90\)) or regression \(R^2 > 0.80\) signals redundancy~\cite{kutner2005applied}.

  (2) Alignment with latent skill space: 
    We assess how well a task is represented by existing factors by: (i) explained variance change when adding the task ($<\!1\%$ suggests redundancy)~\cite{harman1976modern,fabrigar1999evaluating}, (ii) reconstruction error of the task’s scores from the factor model (MSE $>0.10$ indicates novelty)~\cite{jolliffe2002principal}, and (iii) factor loadings (uniformly low $<0.3$ imply poor alignment)~\cite{comrey1992first}. 

    (3) Structural influence:
    We compare PAF solutions with and without the new task. Minimal angular shift ($< 15^\circ$) indicates the structure is unaffected, suggesting redundancy~\citep{bjorck1973numerical, absil2006optimality}.

To compute the \textbf{task novelty-redundancy score}, we thus average the different diagnostics (\(+1\) for novelty, \(-1\) for redundancy) into a continuous score ranging [-1, 1], situating each task on a novel-to-redundant continuum for richer assessment rather than a simple categorical label.

\paragraph{Experimental settings.}
We apply a \emph{leave-one-task-out} procedure, re-fitting the PAF model without each task and recomputing all criteria. This revealed a clear spectrum of task contributions.

At the \textbf{redundant end}, we find tasks like \texttt{glue\_sst2}, \texttt{hellaswag}, and \texttt{imdb}, whose performance is highly predictable from other tasks with minimal structural impact.
At the \textbf{novel end}, we find tasks that inject distinct signal or sharpen underrepresented skill axes. These cluster into themes such as bias probes (\texttt{BLiMP}), safety tests (\texttt{TruthfulQA}), and domain-specific depth (\texttt{GPQA}).

\paragraph{Relation to robustness.}
While in \S\ref{subsec:validRobust} we showed that the eight-factor structure itself is stable, this analysis quantifies how much each 
task sharpens or enriches that structure. 
Novel tasks do not induce new latent dimensions; rather, they contribute distinctive signal within the existing skill space, improving its 
resolution along specific axes.

Alternative diagnostics (e.g., leave-one-skill-out) may further distinguish new skills from noise.

\subsection{How To Efficiently Evaluate New LLMs?}\label{subsec:NewModel}
Evaluating a new LLM typically requires running it across a broad suite of tasks, which is costly and time-consuming. 
Instead, we estimate a model’s full skill profile from a small, diverse task subset.
Our 3-stage pipeline is:

 (1)   Sample a diverse task set:
    We select k=12 tasks (corresponding to \(k\!\approx\!1.5\,C\)~\cite{James2013}) with the highest communalities to maximize latent-skill coverage. The new model is run only on this subset, yielding a performance vector \({p}_{k} \in \mathbb{R}^k\).

    (2) Estimate full task performance:
    Using the \(k\) observed scores and loadings of the chosen tasks \(\Lambda_k \in \mathbb{R}^{k \times C}\) computed based on the training data, we estimate the model’s factor scores \(\hat{\theta} \in \mathbb{R}^C\) via least squares:
        $\hat{\boldsymbol{\theta}} = (\Lambda_k^\top \Lambda_k)^{-1}\Lambda_k^\top \mathbf{p}_k.$
    
    We then reconstruct the model’s full performance with: 
        $\hat{\boldsymbol{\phi}}_{\text{model}} = \Lambda \hat{\boldsymbol{\theta}}.$

    (3) Diagnose model behavior: 
    We assess whether the model conforms to the learned skill space by: 
    (i) applying the Mahalanobis-distance test (\S\ref{subsec:MahalanobisTest}) to detect outliers,
    (ii) evaluating reconstruction error (flagging cases where MSE exceeds \(\mu + \sigma\))~\cite{hawkins1980identification},
    and (iii) analyzing the inferred skill profile \(\hat{\boldsymbol{\theta}}\) for comparative evaluation across models~\citep{thurstone1947multiple}.
This approach yields a low‑cost yet informative 
skill profile. 

\paragraph{Experimental settings.}

To demonstrate this pipeline on unseen models, we perform a \emph{leave-one-model-out} evaluation over 20 held-out LLMs.
Each is evaluated on top-\(k\) tasks, full profile reconstructed, and diagnostics computed. 
All models passed the outlier test (\S\ref{subsec:MahalanobisTest}); only \texttt{bloomz-1b7} exceeded the MSE threshold, indicating a poor reconstruction under the learned skill model.\footnote{\texttt{bloomz-1b7} differs substantially in scale and training regime from the instruction-tuned models used to estimate the latent skill space. As a result, its performance profile lies outside the empirical distribution on which the factor model was learned, where higher reconstruction error is expected.}
Reconstruction error remains low (mean MSE = 0.095), confirming accurate recovery.

We illustrate two characteristic profiles (Figure~\ref{Fig:new_model_projection}:
\texttt{Flan-T5-XL} peaks on Factor~1 (\FactorName{General NLU}) but underperforms on Factors 2, 5, 7, 8 (\FactorName{Entailment, Domain QA, Long-Doc. Comp., Grad-Level Reas.}), matching its C4+Flan training (Fig.~\ref{Fig:t5_xl_factors} in \cref{app:new_model}).  
\texttt{OpenChat-3.5} excels on Factors 1 and 8 but lags on factors 2 and 7, reflecting known RLHF-era tradeoffs (Fig.~\ref{Fig:openchat_factors} in \cref{app:new_model}).
Overall, \texttt{Flan-T5-XL} favors instruction-following with limited domain depth, whereas \texttt{OpenChat} trades fine-grained rigor for recall and academic reasoning.

\subsection{Which Model Is Optimal for my Case?}\label{subsec:chooseModel}
Selecting the optimal model for a new task often demands costly evaluations across many candidates. 
Our goal is not to evaluate a subset of models, but to infer performance across all models from minimal evaluations, enabling full ranking and selection.
Using the same low-rank structure, we instead infer a task’s skill requirements from a small, representative subset of models and predict performance for the rest. 
We proceed in three steps:

  (1)  Sample a diverse model set:
    We select k=12 (\(k\!\approx\!1.5\,C\) \cite{James2013}) models that maximize latent-space coverage via a Max–Min diversity sampler\footnote{greedily maximizes the smallest pairwise distance (often giving better coverage than k‑means centers)}~\citep{BergerWolf2007}. Only these \(k\) models are run on the new task.
    
    Rather than restricting selection to sampled models, we use them to infer the task’s position in latent space and predict performance for all models.

    (2) Estimate task loadings:
    Given a new task with evaluation scores $\mathbf{\phi}_{\text{task}} \in \mathbb{R}^k$ from the $k$ chosen models, and the corresponding factor scores $\hat{\boldsymbol{\Theta}}_k \in \mathbb{R}^{k \times C}$, we estimate the new task's loading vector $\hat{\boldsymbol{\lambda}} \in \mathbb{R}^C$. 

    (3) 
    Predict the rest:
    We predict performances on the new task for all remaining models, rank by the resulting $\hat{\phi}_{\text{task}}$ and select the top candidate.

\paragraph{Experimental settings.}
We use a \emph{leave-one-task-out} protocol across all $B$ tasks.
For each held-out task, we 
fit a PAF model on the remaining $B-1$ tasks
, evaluate \(k\) selected models
, estimate factor loadings, and predict the remaining $M-k$ models. We compute MSE and Pearson \(r\) between predicted and actual scores on the held-out models, and average across tasks.  
This yields a mean normalized MSE below 0.20 and a mean Pearson correlation \(r > 0.85\) for well-represented tasks (e.g., \textsc{GLUE\_RTE}, \textsc{BoolQ}, \textsc{CoQA}); higher errors on underrepresented tasks (e.g., \textsc{bAbI}) highlight gaps in the latent skill map~\citep{Muennighoff2023}.

These results show that performance across models can be accurately predicted from a small subset.

With only \(k\) pilot evaluations and a fixed skill basis, this approach efficiently shortlists strong candidate models for new tasks within or near the existing skill space.\footnote{Our final two applications select small, representative task/model subsets, akin to coreset selection~\citep{Bachem2017coresets,feldman2020introductioncoresetsupdatedsurvey}, but aim at efficient inference in latent skill space rather than geometric approximation.}\footnote{We release code for all applications, 
enabling easy reuse with new benchmarks and models.}

\subsection{Skill Space Maintenance}\label{subsec:maintaining}

\paragraph{Motivation.}
The skill space is learned from a finite set of models and tasks, while the ecosystem evolves rapidly.
A key question is whether a skill space learned on existing models continues to generalize to new model generations, and when it should be updated.

\paragraph{Two regimes.}
In practice, we distinguish between two regimes. 
When new models or tasks lie within the existing skill space, they can be incorporated by projecting them into the learned factor basis.
When diagnostic checks (e.g., reconstruction error or Mahalanobis distance; Stage 3 in Section~\ref{subsec:NewModel}) indicate drift, the factor model can be re-estimated on the expanded performance matrix. 
This requires additional evaluations but remains straightforward and interpretable.

\paragraph{Temporal generalization across model generations.}
We evaluate generalization by training on earlier models ($M_{\text{train}}=48$) and projecting newer models ($M_{\text{test}}=12$) into the learned space.
We measure reconstruction error between predicted and actual performance vectors.

Earlier models are reconstructed accurately (mean normalized MSE $=0.10$), while newer models exhibit moderately higher error (mean MSE in the range $0.32$--$0.42$, depending on the held-out subset size). 
Earlier models are reconstructed accurately (MSE $=0.10$), while newer models ($M_{\text{test}}=8-12$) show moderately higher error (MSE $=0.32$--$0.42$).
Notably, the largest deviations occur for models from the same family (e.g., \texttt{command-a} and \texttt{command-r}), suggesting that architectural or training differences lead to localized shifts in the latent space.

This suggests improvements correspond to movement within an existing capability structure rather than new dimensions.

\paragraph{Takeaway.}

The latent skill space remains largely stable across model generations, with deviations concentrated in a small subset of models.
In practice, this supports a simple operational regime: projection is reliable under moderate distribution shift, while significant increases in reconstruction error or structural diagnostics (Section~\ref{subsec:NewModel}) indicate the need for re-estimation.

These findings suggest a simple maintenance strategy: reuse the factor space when diagnostics remain stable, and trigger re-estimation when reconstruction error or structural checks indicate drift.

\section{Related and Future Work}\label{sec:relatedWork}

Recent work has begun to analyze LLM evaluation benchmarks through latent-variable models, moving beyond aggregate leaderboard averages to uncover shared structure across tasks. 
\citet{burnell2023revealingstructurelanguagemodel} applied maximum-likelihood factor analysis (MLFA) to a small, 
homogeneous set of classification and short-form reasoning tasks, identifying three broad factors interpreted as linguistic understanding, reasoning, and memorization. 
\citet{Ili__2024} extended this psychometric framing to a leaderboard-scale Bayesian factor model, emphasizing a dominant general-ability ($g$) factor with a small number of secondary components.

These studies establish that low-dimensional latent structure exists in task performance, but differ from our work in both scope and emphasis. 
They analyze more limited task collections and focus primarily on identifying broad or hierarchical ability factors, without systematically testing factor stability under task or model perturbations, or exploring downstream evaluation use cases. 
A related line of work models benchmark performance using linear latent-skill assumptions to study scaling laws and extrapolation \citep{polo2025slothscalinglawsllm}. 
While effective for prediction, these approaches assume proportional skill transfer across tasks and do not aim to recover interpretable or stable skill dimensions.

Our work builds on these foundations by applying PAF to a substantially larger and more diverse model–task matrix (60 models × 44 tasks), explicitly isolating shared covariance and validating the resulting latent space under task removal, model subsampling, and dimensionality variation. 
Rather than centering on a single general factor or human cognitive taxonomy, we recover a finer-grained set of stable, data-driven skill dimensions and anchor them to human preferences using Chatbot Arena.

Our recovered skills refine earlier findings: broad dimensions such as `language understanding' or `reasoning' in prior work are decomposed into multiple distinct capabilities (e.g., general NLU, long-doc.\ comp.) that behave differently across tasks, models, and human evaluations.
Where earlier studies establish latent structure, our contribution characterizes it at scale, validates robustness, and supports practical evaluation workflows.

A natural direction is item-level analysis, which may reveal finer-grained skill structure, enable minimal representative subsets, and assess sensitivity to instance selection within benchmarks.
We aim to extend the framework to multilingual and multimodal benchmarks, where skill structure may differ and yield new insights into model capabilities.

\section{Conclusion}\label{sec:conclusion}

This study introduces a psychometric framework for evaluating LLMs, based on factor analysis of model performance across diverse tasks (Fig.~\ref{fig:factor_scores_ours}). By uncovering latent structure in the task performance matrix, our approach identifies eight skills underlying model behavior, ranging from general NLI to instruction-following to domain reasoning. This moves beyond single-score metrics, offering a more compact, interpretable, and multifaceted view of LLM abilities.
These skills remain stable across model and task perturbations, capturing a robust structure that supports consistent evaluation and generalization to new tasks and models.
We develop suite of applications to: assess the novelty of tasks, project new models into the skill space, and efficiently select the best model for a given use case. To support adoption of this methodology, we release all data and code publicly.
Our ultimate goal is to encourage a shift from single-score leaderboards to a more transparent, and practical, skill-based LLM evaluation.

\section*{Acknowledgments}
We thank the anonymous reviewers for their valuable comments. This work has been presented at various seminars and colloquia: the BIU-NLP meeting, Technion CS NLP team meeting, the HUJI NLP Seminar. We thank all participants for comments and fruitful discussion.
This research has been funded by the following funding agencies: a grant from the Israeli Science Foundation (ISF grant 670/23), a KAMIN grant by the Israeli Innovation Authority (IIA),  and a VATAT grant by the the Israeli Planning and Budgeting Committee (PBC), for which we are grateful.

\bibliography{tacl2021}
\bibliographystyle{acl_natbib}

\clearpage
\newpage
\appendix
\section{principal Axis Factoring} \label{app:paf}

In PAF, we assume the latent‑variable factorization
\[
  \PerformanceMat \;=\; \ThetaMat\LambdaMat^{\!\top} + \epsilonMat,
  \tag{\ref{eq:paf-matrix}}
\]
so that each row satisfies
\(\mathbf p_i = \bm\theta_i^{\top}\LambdaMat^{\!\top} + \bm\varepsilon_i\).

We further assume the following statistical conditions:
\[
\begin{aligned}
\mathbb E[\bm\theta_i] &= \mathbf 0                &\qquad
\operatorname{Cov}(\bm\theta_i)            &= \mathbf I_{C} \\[2pt]
\mathbb E[\bm\varepsilon_i] &= \mathbf 0            &\qquad
\operatorname{Cov}(\bm\varepsilon_i)       &= \boldsymbol{\Psi} \\[2pt]
\boldsymbol{\Psi}          &= \operatorname{diag}(\psi_1,\dots,\psi_B) &\qquad
\operatorname{Cov}(\bm\theta_i,\bm\varepsilon_i) &= \mathbf 0
\end{aligned}
\]

The population covariance (or correlation) matrix of task scores is
decomposed as
\[
  \mathbf R
  \;=\;
  \LambdaMat\LambdaMat^{\!\top} + \boldsymbol{\Psi},
\]
where \(\LambdaMat\LambdaMat^{\!\top}\) captures the \emph{shared} variance
explained by the latent skills and
\(\boldsymbol{\Psi}=\operatorname{diag}(\psi_1,\dots,\psi_B)\)
collects the \emph{unique} (task‑specific) variances.

\paragraph{Relation to principal–component analysis (PCA)}
PCA factorizes the same matrix as
\(\mathbf R=\mathbf U\mathbf D\mathbf U^{\!\top}\) with
orthonormal loadings \(\mathbf U\) and diagonal eigenvalues \(\mathbf D\),
seeking directions that maximise total variance.  
Unlike FA, PCA
\emph{(i)} does not separate shared from unique variance,  
\emph{(ii)} requires orthogonality of the components, and  
\emph{(iii)} identifies a unique loading matrix up to sign.
FA, by contrast, models only the off‑diagonal structure
and therefore allows rotational indeterminacy:
any nonsingular \(\mathbf T\) with
\(\boldsymbol{\Lambda}\mathbf T\) spans the same latent subspace.

In FA the loading matrix $\boldsymbol{\Lambda}$ is not
identifiable up to \emph{any} non‑singular rotation
$(\boldsymbol{\Lambda},\mathbf f)\!\to\!(\boldsymbol{\Lambda}\mathbf T,
\mathbf T^{-1}\mathbf f)$, so demanding
$\boldsymbol{\Lambda}^{\!\top}\boldsymbol{\Lambda}=\mathbf I_m$ would merely
freeze one arbitrary rotation without improving model fit. PAF aims to determine the \emph{subspace} spanned by the latent loadings. PCA identifies each loading with a concrete direction that maximizes total variance, and rotating that direction will violate optimality. Orthogonality in
the ordinary dot‑product is also conceptually mismatched: each variable is
weighted by its reliability $1/\psi_j$, making the natural metric
$\langle\mathbf a,\mathbf b\rangle_{\Psi^{-1}}
=\mathbf a^{\!\top}\boldsymbol{\Psi}^{-1}\mathbf b$, not the Euclidean one.
PAF allows non‑orthogonal loadings, and later apply Varimax or Promax
rotations, which maximize sparsity and yield factors that align with
interpretable domain concepts.

\subsubsection*{Iterative solution algorithm for PAF}
PAF estimates the loading matrix \(\LambdaMat\) and uniquenesses
\(\boldsymbol{\Psi}\) by minimizing
\(\|\mathbf R-\LambdaMat\LambdaMat^{\!\top}-\boldsymbol{\Psi}\|_F^{2}\).
The key quantity is the \emph{communality}
\(h_j^{2}=\sum_{c=1}^{C}\lambda_{jc}^{2}\), the portion of task
\(j\)’s variance explained by the common factors.  
The procedure alternates between (i) recomputing \(\LambdaMat\) from a
reduced correlation matrix whose diagonal equals the current
communalities, and (ii) updating the communalities from the new
loadings, until successive updates differ by less than a tolerance
\(\varepsilon\).
\begin{enumerate}[leftmargin=*]
\item \textbf{Initial communalities}  
      Set \(h_j^{2,(0)}\) to the squared multiple correlation of task
      \(j\) with all others (or simply 1).

\item \textbf{Reduced correlation matrix}  
      Replace the diagonal of \(\mathbf R\) with the current communalities
      to obtain \(\mathbf R^{(t)}\).

\item \textbf{SVD / eigen‑step (loadings update)}  
      Compute the truncated eigen‑decomposition  
      \[
         \mathbf R^{(t)} = \mathbf Q_c\,\boldsymbol{\Delta}_c\,\mathbf Q_c^{\!\top},
      \]
      where \(\mathbf Q_c\in\mathbb R^{B\times C}\) contains the
      eigenvectors associated with the \(C\) largest eigenvalues, and
      \(\boldsymbol{\Delta}_c=\operatorname{diag}(\delta_1,\dots,\delta_C)\).
      The updated loading matrix is
      \[
         \LambdaMat^{(t+1)} \;=\; \mathbf Q_c\,\boldsymbol{\Delta}_c^{1/2},
      \]
      i.e.\ each retained eigenvector is scaled by the square root of its
      eigenvalue.

\item \textbf{Update communalities}  
      \(h_j^{2,(t+1)}=\sum_{c=1}^C\lambda_{jc}^{2,(t+1)}\).

\item \textbf{Convergence check}  
      Stop when \(\max_j|h_j^{2,(t+1)}-h_j^{2,(t)}|<\varepsilon\)
      (e.g.\ \(\varepsilon=10^{-4}\)); otherwise iterate.
\end{enumerate}

After convergence, set
\(\boldsymbol{\Psi}=\operatorname{diag}(1-h_1^{2,*},\dots,1-h_B^{2,*})\).
Because \(\LambdaMat\) is only unique up to rotation, an
orthogonal (e.g.\ Varimax) or oblique (e.g.\ Promax) rotation can be
applied to \(\LambdaMat^{*}\) to enhance interpretability without altering
model fit.

\subsubsection*{Rotating the loading matrix}
After estimating an initial (possibly non‑sparse) loading matrix,
researchers often apply an additional rotation to enhance
interpretability.

\begin{itemize}
  \item \textbf{Orthogonal rotations} preserve factor independence.  
        The Orthomax family maximises  
        \(\displaystyle\sum_{c=1}^C\bigl[\sum_{j=1}^B\lambda_{jc}^4
        -\gamma B^{-1}(\sum_{j=1}^B\lambda_{jc}^2)^2\bigr]\).
        Common choices:\\
        \hspace*{1em}• \emph{Varimax} (\(\gamma\!=\!1\)): encourages each factor to load
        strongly on a small subset of tasks;\\
        \hspace*{1em}• \emph{Quartimax} (\(\gamma\!=\!0\)): encourages each task
        to load on as few factors as possible;\\
        \hspace*{1em}• \emph{Equamax} (\(\gamma\!=\!C/2\)).
  \item \textbf{Oblique rotations} allow correlated factors.  
        Promax starts from a Varimax solution, then “bends” each column
        by raising its entries to a power (\(p\!\approx\!3\text{–}4\))
        and re‑orthogonalising, yielding a sparse but correlated
        loading pattern.  
        Other popular oblique criteria include \emph{Direct Oblimin}
        and \emph{Geomin}.
\end{itemize}

\subsubsection*{Assessing the novelty of a candidate task}

Suppose a new, $z$‑scored task vector 
$\mathbf p_{\text{cand}}\!\in\!\mathbb R^{M}$ (scores of the $M$ models) 
is proposed. With the skill matrix 
$\ThetaMat\!\in\!\mathbb R^{M\times C}$ and the fitted loadings
$\LambdaMat\!\in\!\mathbb R^{B\times C}$ fixed, we evaluate whether the
candidate adds \emph{shared} information that is not already captured by the
existing $B$ rows of~$\LambdaMat$.

\paragraph{1.\  Project the candidate into skill space}  
Obtain its loading vector by least–squares regression on the factor scores
(the analogue of \cref{eq:thomson-BLP}):
\[
  \bm\lambda_{\text{cand}}
  = (\ThetaMat^{\!\top}\ThetaMat)^{-1}\ThetaMat^{\!\top}\mathbf p_{\text{cand}}
  \quad\in\mathbb R^{C}.
\]

\paragraph{2.\  Compute communality and uniqueness}  
\[
  h_{\text{cand}}^{2} = \|\bm\lambda_{\text{cand}}\|_{2}^{2},
  \qquad
  \psi_{\text{cand}}   = 1-h_{\text{cand}}^{2}.
\]
A high $h_{\text{cand}}^{2}$ indicates the task is well explained by the
latent skills; a high $\psi_{\text{cand}}$ signals mostly idiosyncratic
variance.

\paragraph{3.\  Compare loading profiles}  
Measure cosine similarity in skills space:
\[
  \rho_{j} \;=\;
  \frac{\bm\lambda_{\text{cand}}^{\!\top}\bm\lambda_{j}}
       {\|\bm\lambda_{\text{cand}}\|_{2}\,\|\bm\lambda_{j}\|_{2}},
  \quad j=1,\dots,B.
\]
If $\max_{j}\rho_{j}$ exceeds a threshold (e.g.\ $0.9$) the candidate’s
pattern is essentially a duplicate of task $j$.

\paragraph{4.\  Residual correlation check}  
Compute the residual vector
$\mathbf r = \mathbf p_{\text{cand}} - \ThetaMat\bm\lambda_{\text{cand}}$ and
its correlation with each existing task’s residual.  Large residual
correlations ($|\operatorname{corr}(r,\varepsilon_j)|>0.2$) suggest that the
current factor structure would need to expand to accommodate truly new shared
variance.

\section{Skill Naming and Interpretation Procedure}\label{app:skillNaming}

This appendix describes the procedure used to assign descriptive labels to the latent factors extracted by PAF. The goal of this two-step process is to provide concise, interpretable summaries of each factor’s semantic content, without affecting the underlying factor structure or model scores.

After PAF produces $C$ latent factors, we treat each as a skill which models vary on, and assign to it a concise skill label.
\paragraph{Step 1 - Selecting representative tasks.}
For each latent factor $c$, we identify the tasks with the highest absolute loadings (|$\lambda_{dc}|$), by computing the z-scored absolute loadings \footnote{\(z_{dc}=(|\lambda_{dc}|-\mu_c)/\sigma_c\), each loading’s deviation from its factor’s mean in STD units.} and selecting the tasks where \(z\ge1\) ($\approx$ top 16\%).
Tasks with \(\lambda_{dc} > 0\) are assigned to the positive set; those with \(\lambda_{dc} < 0\) are assigned to the negative set.

\paragraph{Step 2 - LLM-assisted labeling.}

Each factor is labeled by prompting an LLM with short descriptions of its representative tasks identified in Step~1. The model is instructed to propose a concise descriptive name that summarizes the shared capability reflected by these tasks. The prompt does not restrict the label to predefined categories, allowing labels to emerge naturally from task semantics.
The authors review and refine the suggested labels for clarity and consistency.

In order to capture what each factor name is as a skill we had the next prompt. 
``We produced a PAF model where the original data matrix contains the performances of LLMs on a set of tasks (LLMs as the observations and tasks are the variables). The PAF model produced 8 factors we treat as LLM skills. In order to give a skill name to each factor - attached are the strongest loadings for each one of the factors. 
Additionally, here is a description of each one of the tasks that were evaluated. 
Can you please take into account both the positive and negative loadings and label each one of the factors as a skill?''
\footnote{We verified that varying the labeling prompt (e.g., per-factor vs. all-factors, with or without the word `skill') produced identical factor interpretations, confirming that the semantic consistency arises from the loadings themselves rather than prompt phrasing.}
In Table~\ref{tab:all_tasks_merged} are the different tasks descriptions. 

\begin{table*}[t]
\centering
\small
\scriptsize                  
\begin{tabular}{@{}>{\raggedright\arraybackslash}p{4cm}
                >{\raggedright\arraybackslash}p{2cm}
                >{\raggedright\arraybackslash}p{1.2cm}
                >{\raggedright\arraybackslash}p{7cm}@{}}
\toprule
\textbf{Task} & \textbf{Output Type} & \textbf{Eval Metric} & \textbf{Short Description}\\
\midrule
GLUE – CoLA \cite{wang2019gluemultitaskbenchmarkanalysis} & binary class. & F1 & Acceptability judgments for grammatical well-formedness.\\
GLUE – MNLI \cite{wang2019gluemultitaskbenchmarkanalysis} & mult. choice  & F1 & 3-way natural-language inference across genres.\\
GLUE – MRPC \cite{wang2019gluemultitaskbenchmarkanalysis} & binary class. & F1 & Sentence-level paraphrase detection for news.\\
GLUE – QNLI \cite{wang2019gluemultitaskbenchmarkanalysis} & binary class. & F1 & Sentence-question entailment derived from SQuAD.\\
GLUE – QQP  \cite{wang2019gluemultitaskbenchmarkanalysis} & binary class. & F1 & Paraphrase detection for Quora question pairs.\\
GLUE – RTE  \cite{wang2019gluemultitaskbenchmarkanalysis} & binary class. & F1 & Textual-entailment classification from PASCAL challenges.\\
GLUE – SST-2 \cite{wang2019gluemultitaskbenchmarkanalysis} & binary class. & F1 & Sentiment analysis of movie snippets.\\
GLUE – STS-B \cite{wang2019gluemultitaskbenchmarkanalysis} & regression & F1 & Continuous sentence-pair semantic similarity.\\
GLUE – WNLI \cite{wang2019gluemultitaskbenchmarkanalysis} & binary class. & F1 & Coreference-switch paraphrase task with high variance.\\
MMLU \cite{hendrycks2021measuringmassivemultitasklanguage} & mult. choice & Acc. & 57-subject multi-choice exams measuring broad knowledge.\\
SQuAD \cite{rajpurkar2016squad100000questionsmachine} & span extract. & LLM & Answer extraction from paragraphs.\\
m3exam \cite{zhang2023m3exammultilingualmultimodalmultilevel} & mult. choice & Acc. & Multilingual, multimodal exam-style benchmark.\\
MedQA \cite{jin2020diseasedoespatienthave} & mult. choice & LLM & USMLE-style medical questions.\\
GSM8K \cite{cobbe2021gsm8k} & generation & LLM & Grade-school math word problems with numeric answers.\\
Math \cite{hendrycks2021measuringmathematicalproblemsolving} & generation & LLM & Diverse mathematics problem solving.\\
LegalBench (MC) \cite{guha2023legalbenchcollaborativelybuiltbenchmark} & mult. choice & Acc. & Suite of specialised legal-reasoning tasks.\\
LegalBench (Gen) \cite{guha2023legalbenchcollaborativelybuiltbenchmark} & generation & LLM & Generation-style subsets of LegalBench.\\
BoolQ \cite{clark2019boolqexploringsurprisingdifficulty} & binary class. & F1 & Yes/No QA over Wikipedia passages.\\
HellaSwag \cite{zellers2019hellaswagmachinereallyfinish} & mult. choice & Acc. & Commonsense sentence completion with adversarial distractors.\\
OpenBookQA \cite{mihaylov2018suitarmorconductelectricity} & mult. choice & LLM & Elementary-science QA with a supplied fact and commonsense.\\
TruthfulQA (Gen) \cite{lin2022truthfulqameasuringmodelsmimic} & generation & LLM & Generation benchmark rewarding factual truth over misconceptions.\\
TruthfulQA (MC) \cite{lin2022truthfulqameasuringmodelsmimic} & mult. choice & LLM & Multiple-choice version of TruthfulQA.\\
MS~MARCO \cite{bajaj2018msmarcohumangenerated} & generation & LLM & Passage retrieval / ranking for web queries.\\
TriviaQA \cite{joshi2017triviaqalargescaledistantly} & span extract. & LLM & Open-domain trivia QA over large corpora.\\
DailyDialog \cite{li2017dailydialogmanuallylabelledmultiturn} & mult. choice & LLM & Dialogue-act / emotion classification in informal chats.\\
HotpotQA \cite{yang2018hotpotqadatasetdiverseexplainable} & generation & LLM & Multi-hop Wikipedia QA requiring two supporting docs.\\
CoQA \cite{reddy2019coqaconversationalquestionanswering} & generation & LLM & Conversational QA grounded in documents.\\
BIG-Bench (MC) \cite{srivastava2023imitationgamequantifyingextrapolating} & mult. choice & F1 & Mixed-format challenge stressing broad reasoning.\\
BIG-Bench (Gen) \cite{srivastava2023imitationgamequantifyingextrapolating} & generation & LLM & Generation subsets of BIG-Bench.\\
QuAC \cite{choi2018quacquestionanswering} & generation & LLM & Multi-turn dialogical QA over Wikipedia.\\
CNN/DailyMail \cite{see-etal-2017-get} & generation & LLM & Abstractive news-article summarisation.\\
XSum \cite{narayan2018dontdetailsjustsummary} & generation & LLM & One-sentence abstractive summaries of BBC news.\\
IMDB \cite{maas-EtAl:2011:ACL-HLT2011} & binary class. & F1 & Sentiment of full movie reviews.\\
bAbI \cite{dodge2016evaluating} & generation & LLM & Synthetic stories with rule-based multi-step queries.\\
BLiMP \cite{warstadt2023blimpbenchmarklinguisticminimal} & binary class. & Acc. & 67 minimal-pair syntactic phenomena for grammaticality.\\
BBQ \cite{parrish2022bbqhandbuiltbiasbenchmark} & mult. choice & Acc. & Social-bias sensitivity under controlled knowledge settings.\\
ARB \cite{sawada2023arbadvancedreasoningbenchmark} & generation & F1 & Adversarial cross-domain questions to fool LLMs.\\
Legal-Support \cite{lighteval} & binary class. & F1 & Statute sentences supporting/contradicting a claim.\\
Synthetic-Reasoning \cite{lighteval} & generation & LLM & GPT-generated science problems requiring chain-of-thought.\\
DocNLI \cite{yin2021docnlilargescaledatasetdocumentlevel} & binary class. & F1 & Document-level natural-language inference.\\
Contract-NLI \cite{koreeda2021contractnlidatasetdocumentlevelnatural} & mult. choice & F1 & Entail/contradict/neutral classification on contracts.\\
Quality \cite{pang2022qualityquestionansweringlong} & mult. choice & F1 & Long-document QA measuring factuality and completeness.\\
QMSum \cite{zhong2021qmsumnewbenchmarkquerybased} & generation & LLM & Query-focused meeting summarisation.\\
CivilComments \cite{borkan2019nuancedmetricsmeasuringunintended} & regression & F1 & Toxicity classification of online comments.\\
GPQA \cite{rein2023gpqagraduatelevelgoogleproofqa} & generation & LLM & Graduate-level physics questions with diagrams.\\
SQuALity \cite{wang2022squalitybuildinglongdocumentsummarization} & generation & LLM & Document-level summarisation graded for fidelity and coherence.\\
\bottomrule
\end{tabular}
\caption{Tasks used in the factor analysis, showing output type, common evaluation metrics, the metric adopted in this study, and a concise task description.}
\label{tab:all_tasks_merged}
\end{table*}

\paragraph{Interpretation scope.}
The descriptive names assigned to each latent skill (e.g., ``Precision \& Fidelity'') are interpretive summaries based on the highest-loading tasks.
These labels aid exposition but do not affect the underlying factor structure, which is objectively derived from the shared-variance model.
The assigned skill names are descriptive summaries intended to aid interpretation; while alternative labels are possible, the underlying factor structure and model scores are unaffected.

The full prompt template and implementation details are provided in \cref{app:implementationDetails}.

\section{Implementation Details}\label{app:implementationDetails}

\paragraph{Factor Analysis.}  
We fit a PAF model using the \texttt{factor\_analyzer} Python package, extracting $C=8$ factors with \texttt{rotation=``varimax''}. The number of factors was selected based on eigenvalue thresholds and explained variance inspection.

\paragraph{Factor Scores.}  
Model-level factor scores were computed using the regression method from \texttt{factor\_analyzer}. These scores represent each model’s strength on every latent skill and are used throughout all downstream applications (e.g., novelty detection, model profiling, skill-aware evaluation).

\paragraph{Factor Labeling.}  
To label each factor as a latent skill, we passed the high-loading task sets through two LLMs—\texttt{deepseek-R1} and \texttt{gpt-o3-pro}—prompted to propose intuitive skill names based on shared task properties. Labels were selected via manual refinement.

\section{What Drives Model Performance Across Skills}  \label{app:ModelsSkillsAnalysis}
Our analysis reveals a sharp dichotomy in what drives model performance across the eight skills. 
Seven of the eight factors reward \emph{specialisation, not size}: the top models are 7–13 B models whose training curriculum—pre-training data, fine-tune tasks, and even architecture—is precisely aligned with the skill being measured.
Only Factor 4 (Truthfulness \& Instruction-Following) shows a noticeable size correlation ($\rho = 0.72$), driven mainly by the Gemini-1.5-Pro. 

Below, we detail how each skill aligns with specific data and tuning choices.

\begin{itemize}
  \item \textbf{Factor~1 \,(General Language Understanding).} 
  Most models cluster near the mean ($\pm1,\sigma$), confirming broad saturation of this fundamental skill. However, models like \textit{Flan-T5-XL}, \textit{Flan-UL2}, and \textit{Command-R-A-03-2025} gain an edge through intensive fine-tuning on massive datasets of short classification prompts (e.g., sentiment analysis, textual entailment). This specialization in precise linguistic tasks—disambiguating syntax, resolving references, and mapping instructions to exact outputs—captures the remaining performance margin that generic pre-training misses.

  \item \textbf{Factor~2 \,(Fine-grained Entailment \& Bias Reasoning).} \emph{Falcon-7B-Instruct} and \emph{Koala-7B/13B} excel because their chat mixtures \emph{over-sample sentence-pair entailment and social-bias prompts}, mirroring QNLI, RTE, BBQ and BLiMP.

  \item \textbf{Factor~3 \,(Document Reading, Retrieval \& Summarization).} 
  \emph{Gemini-1.5-Pro} scores highest, closely followed by the seq-to-seq \emph{T0/T0p/T0pp} family. 
  All were tuned on HotpotQA, MS~MARCO, SQuAD and XSum, and use retrieval-augmented prompting (Gemini) or encoder–decoder architecture (T0 variants) that excels at ``extract then rewrite''.
  This direct match between training mix, model design and the factor’s long-context tasks explains their advantage.
  
  \item \textbf{Factor~4 \, (Truthfulness \& Instruction Following Generation}
  The sole standout is \emph{Gemini-1.5-Pro} ($+5.2\,\sigma$), pairing large capacity with long RLHF/RLAIF on factual tasks.
  Other large models cluster near the mean. A midsize tier that include dedicated truthfulness tuning (e.g. \emph{Command‑A‑03‑2025}, \emph{GPT‑4o‑Mini}, \emph{Dolly‑v2‑3B}, and the \emph{Qwen 2.5+} instruction family) rank next. While models of similar size trained only on generic instruction corpora (e.g., \emph{Dolly‑v2‑7 B/12 B}) lag well below. The factor therefore rewards capacity amplified by targeted factual‑accuracy supervision rather than scale alone.

  \item \textbf{Factor~5 \,(Specialized Domain-Knowledge QA).} 
  \emph{GPT-4o-Mini} ranks first, with \emph{Meta-Llama-3.1-8B-Instruct} close behind. Both are fine-tuned on high-quality synthetic medical, legal and technical Q\&A generated by larger parent models (GPT-4o and Meta-405B), showing that targeted, domain-specific supervision—not scale alone—drives this skill.
  
  \item \textbf{Factor~6 \,(Social \& Ethical Judgment).} 
   \emph{GPT-4o-Mini} ranks first, followed by \emph{BLOOMZ-3B} and \emph{BLOOMZ-1B7}. Their edge comes from safety-focused supervision—multilingual xP3 toxicity prompts or red-team (for all three) + RLHF from human feedback pipeline (for gpt-4o-mini)—that trains strong negative constraints. Models lacking such tuning, regardless of size, cluster near the mean.

  \item \textbf{Factor~7 \,(Faithful Summarization \& Quantitative Precision).} 
  \emph{Phi-4} leads ($+2.5\,\sigma$), with \emph{DeepSeek-Distill-Llama-8 B} and \emph{Qwen 2.5-7 B-Instr} close behind ($> +2\,\sigma$).
  All three—and, in fact, every model scoring above ($+1.5\,\sigma$)—were trained on ``textbook‑quality'' synthetic math/logic chains that reward token‑level exactness (generated by GPT‑4, DeepSeek‑R1, and a Qwen‑Max + GPT‑4 mix, respectively). These curated, step‑by‑step reasoning traces—not raw parameter count—are what drive this skill.

  \item \textbf{Factor~8 \,(Graduate-Level Scientific and Legal Reasoning).} 
  \emph{Pythia-Chat-7B} and \emph{Claude-3.5-Haiku} (both $\sim$7\,B) top the column, beating much larger Gemma and Qwen variants.
  Both combine arXiv-heavy pre-training with a second-stage finetune on graduate-level STEM QA sets (ARC-Ch, PubMed-QA, MMLU-STEM), showing that targeted science instruction—not scale—drives this skill.
\end{itemize}

\medskip
\textbf{Take-away.} 
Except for Factor 4, adding parameters delivers little; High scores instead reflect alignment between training data, model architecture, and the skill being measured.
Including more \(\ge\!70\) B models might steepen the size slope for Factor 4 but is unlikely to alter the weak size influence elsewhere.

\section{Emergent Generalizations in Factor Structure}  \label{app:skillsGeneralization}

Our FA reveals five structural regularities that clarify how latent skills organize and relate to model behavior:

\subsection{Task Output Format Influence Factor Membership.}  
For most factors, salient tasks ($|\lambda|\ge0.50$) cluster by output format: Classification tasks (e.g., GLUE, BBQ, BLiMP) load on Factors~1, 2, and~6, while generation tasks (e.g., SQuAD, XSum, Math) cluster in Factors~3, 4, and~7. Factors~5 and~8 mix formats, suggesting that content and difficulty sometimes override format.

\subsection{Non-Saturated Tasks Shape the Factor Space.}  
Tasks on which models show wide performance gaps—that is, those with large score variance—contribute most to factor formation.
Tasks with wide model score variance shape factor structure most strongly. Easy, saturated tasks (e.g., SST-2, MNLI) compress into general Factor~1, while harder tasks (e.g., MedQA, LegalBench) form distinct axes.
Model diversity—across size, architecture, and training—is crucial to reveal them.

\subsection{Task Difficulty Overrides Domain.}  \label{par:difficultyDomain}
Task complexity—not domain—often dictates factor placement. Simple domain-specific tasks (e.g., Legal‑Support, GSM8K) fall under general NLU (Factor~1), while harder variants (e.g., LegalBench, SQuALity, BoolQ) define specialized dimensions (Factors~5 and ~7). Behavioral demands, not surface labels, drive separation.

\subsection{Instruction-Following Forms a Distinct Skill Axis.}  
Factor~4 captures models’ ability to follow detailed instructions, with high loadings from \textsc{TruthfulQA} and \textsc{SQuALity}. Models like \texttt{Flan-T5} and \texttt{OpenChat-3.5} excel here, reflecting benefits from alignment tuning.

\subsection{Token-Level Precision Forms a Cross-Domain Factor.} \label{par:tokenLevelPrecision} 
Factor~7 captures tasks where small errors break performance—e.g., \textsc{Math}, \textsc{SQuality}—regardless of domain. These tasks demand token-level \emph{exactness} in reasoning or generation. Top models (e.g., \texttt{command-r}, \texttt{bloomz-3b}) are all instruction-tuned, linking this token-level fidelity skill to alignment.

\textit{Together}, Factors~4 and~7 reveal distinct outcomes of instruction tuning: it builds a dedicated instruction-following skill (Factor~4) and separately boosts token-level precision on exacting tasks (Factor~7). These aligned capabilities emerge beyond what traditional NLU datasets capture.

\section{List of Tasks} \label{App:fullDataset}
Table~\ref{tab:all_tasks_merged} list the full list of tasks which we used to evaluate the models upon as well as the type of the task: i.e. binary classification, multiple choice classification or generation. 
LLM indicate LLM as a judge method \cite{zheng2023judgingllmasajudgemtbenchchatbot}.

\section{Instruction per Task}\label{App:BenchmarkInstruction}
In this section, we elaborate the algorithm for constructing the instructions that we used for each one of the tasks on all models. 

We adopted a systematic, iterative method for instruction creation. 
First, a base instruction was generated using Gemini-pro-1.5 \citep{geminiteam2024gemini15unlockingmultimodal} and evaluated on a subset of 50 labeled examples across 10 diverse models. An instruction was accepted for a model if at least 70\% of outputs were valid. If not, we refined the instruction using targeted prompts to rephrase it for better clarity, and re-evaluated it across all models to maintain broad applicability. This iterative refinement continued until the instruction achieved at least 70\% valid outputs for all 10 models. Subsequently, the instruction was further validated on additional models, retaining only those models with valid outputs for at least 50\% of the examples. Our algorithm for creating such instructions are detailed in Algorithm~\ref{algo:creatingInstruction}.

\begin{algorithm}[t] 
\caption{Iterative Instruction Refinement}\label{algo:creatingInstruction}
\begin{algorithmic}[1]
\Require Dataset $D$, base instruction $I_0$, models $M$, subset $M_s = \{M_1, \dots, M_{10}\} \subset M$, subset $S \subset D$, threshold $\tau = 70\%$. PROMPT(I,S) = rephrasing prompt
\Ensure Refined instruction $I^*$, final model list $M_{\text{final}} \subseteq M$.

\State Initialize $I \leftarrow I_0$, $M_{\text{final}} \leftarrow \emptyset$, randomize $M_s$.
\State Call \textsc{RefineInstruction}($I$, $M_s$, $S$, $\tau$).
\State Validate $I$ on additional models; include in $M_{\text{final}}$ if $V_i \geq 50\%$.
\State \textbf{Return} $I^* \leftarrow I$, $M_{\text{final}}$.
\Procedure{\textsc{RefineInstruction}}{$I$, $M_s$, $S$, $\tau$}
    \For{each $M_i \in M_s$}
        \State Evaluate $S$ using $I$ on $M_i$; compute validity rate $V_i$.
        \If{$V_i < \tau$}
            \State Refine $I$ using PROMPT(I,S); update $I \leftarrow I'$.
            \For{each $M_j \in M_{\text{final}}$}
                \State Re-evaluate $S$ using $I$ on $M_j$; update $V_j$.
                \If{$V_j < \tau$} 
                    \State Call \textsc{RefineInstruction}($I$, $M_s$, $S$, $\tau$) \Comment{Recursive step}
                \EndIf
            \EndFor
            \State Call \textsc{RefineInstruction}($I$, $M_s$, $S$, $\tau$) \Comment{Recursive step}
        \Else
            \State Add $M_i$ to $M_{\text{final}}$.
        \EndIf
    \EndFor
\EndProcedure
\end{algorithmic}
\end{algorithm}

\section{Robustness, Factor Diagnostics}\label{sub:resAttRobust}

\begin{figure}[t]
\centering
\includegraphics[width=0.4\textwidth]{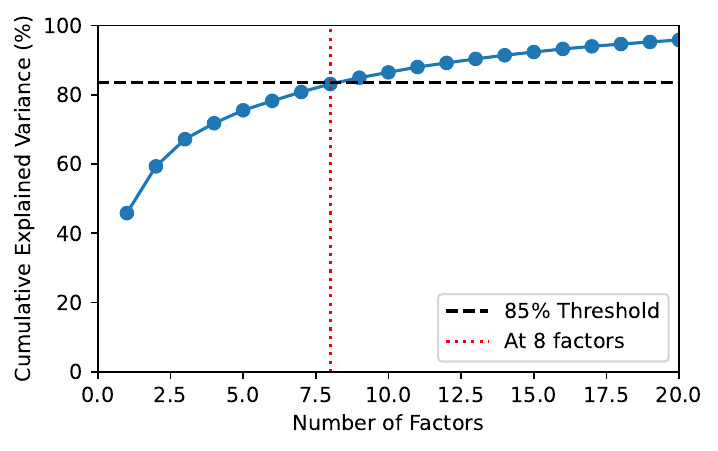}
\caption{Cumulative explained variance to choose optimal number of factors.}
\label{Fig:fa_n_factors_explained}
\end{figure}

\paragraph{Number of factors.}\label{subapp:numFactors}
In Fig.~\ref{Fig:fa_n_factors_explained} is the cumulative explained variance that we used to detect the optimal number of latent factors. 

\paragraph{Uniqueness.}\label{subapp:uniqueness}

\begin{figure}[t]
\centering
\includegraphics[width=0.5\textwidth]{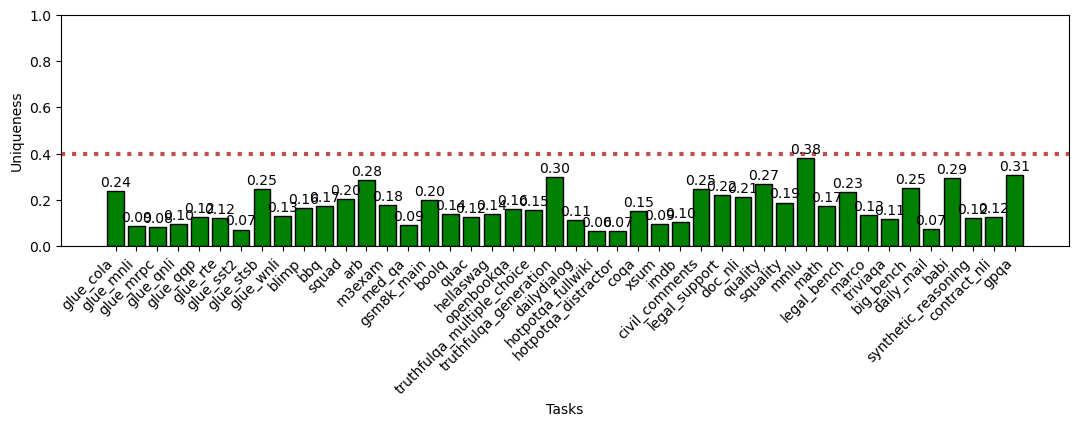}
\caption{Uniqueness of each task}
\label{Fig:uniqueness}
\end{figure}

In Fig.~\ref{Fig:uniqueness} we present the uniqueness, which quantifies how much of a task's variance is not captured by the latent factors. 

\begin{figure*}[t]
\centering
\begin{subfigure}[b]{0.48\textwidth}
    \centering
    \includegraphics[width=0.8\linewidth]{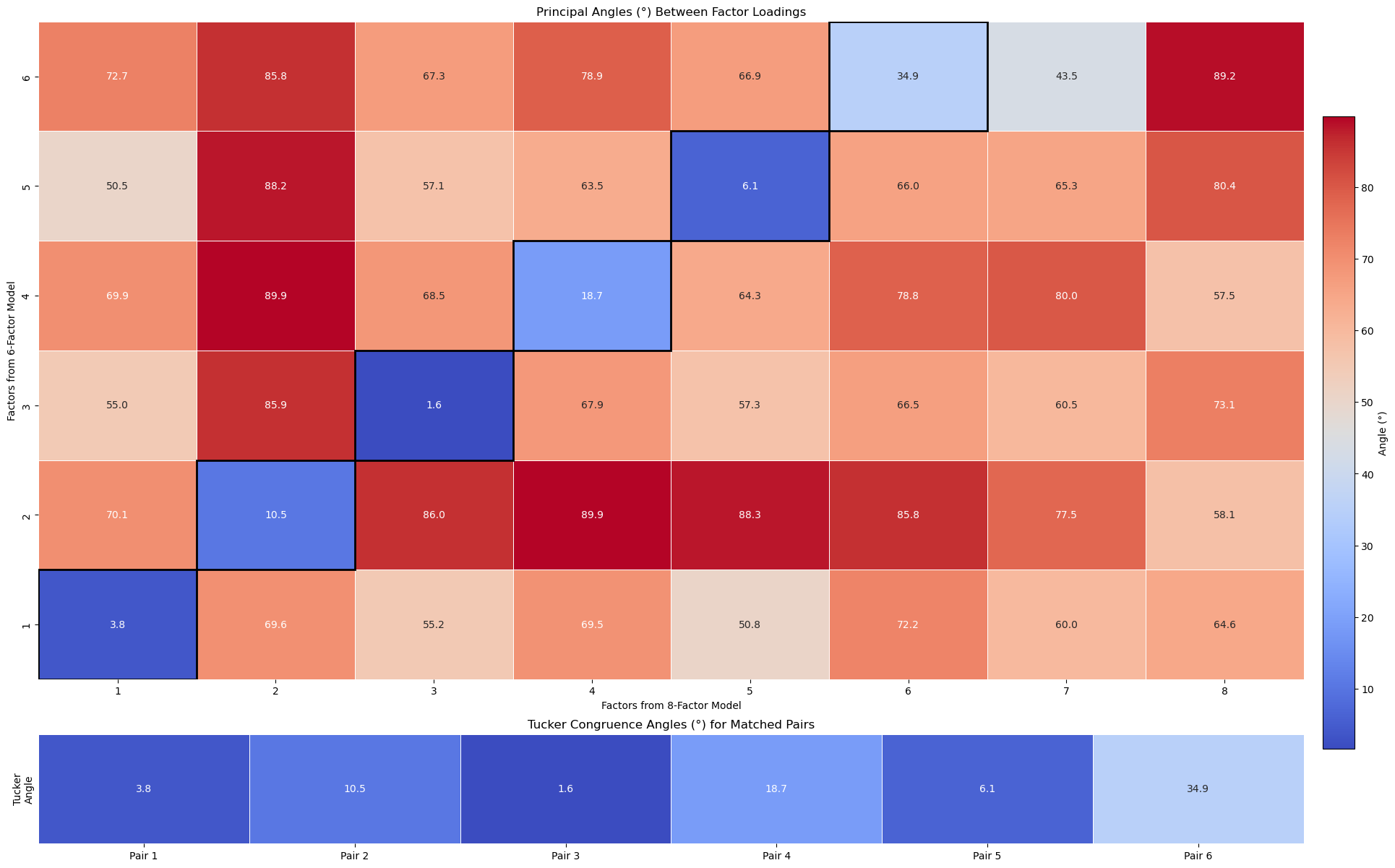}
    \caption{6 vs 8 factors}
    \label{Fig:angle_6_8}
\end{subfigure}
\hfill
\begin{subfigure}[b]{0.48\textwidth}
    \centering
    \includegraphics[width=0.8\linewidth]{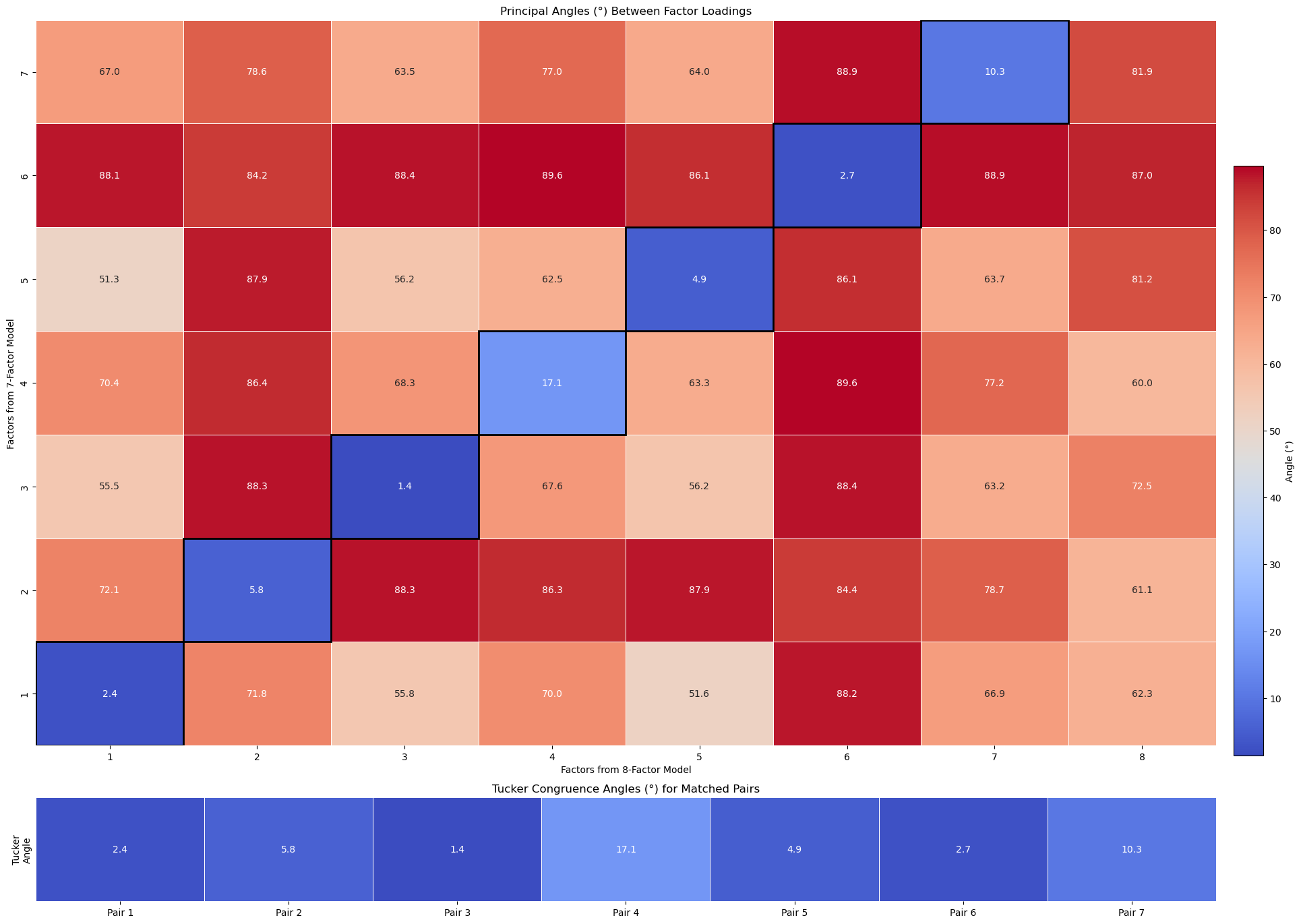}
    \caption{7 vs 8 factors}
    \label{Fig:angle_7_8}
\end{subfigure}

\begin{subfigure}[b]{0.48\textwidth}
    \centering
    \includegraphics[width=0.8\linewidth]{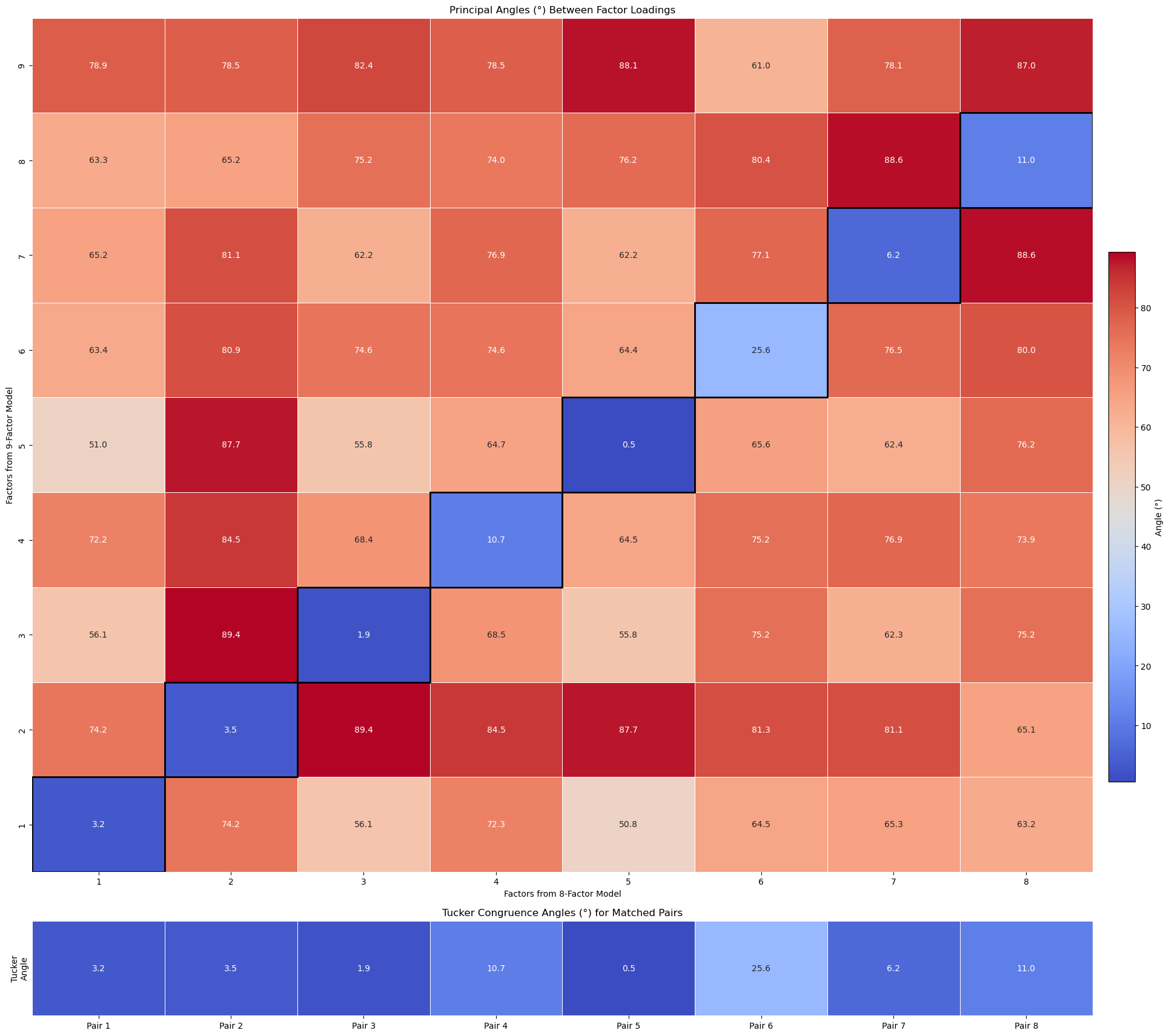}
    \caption{8 vs 9 factors}
    \label{Fig:angle_8_9}
\end{subfigure}
\hfill
\begin{subfigure}[b]{0.48\textwidth}
    \centering
    \includegraphics[width=0.8\linewidth]{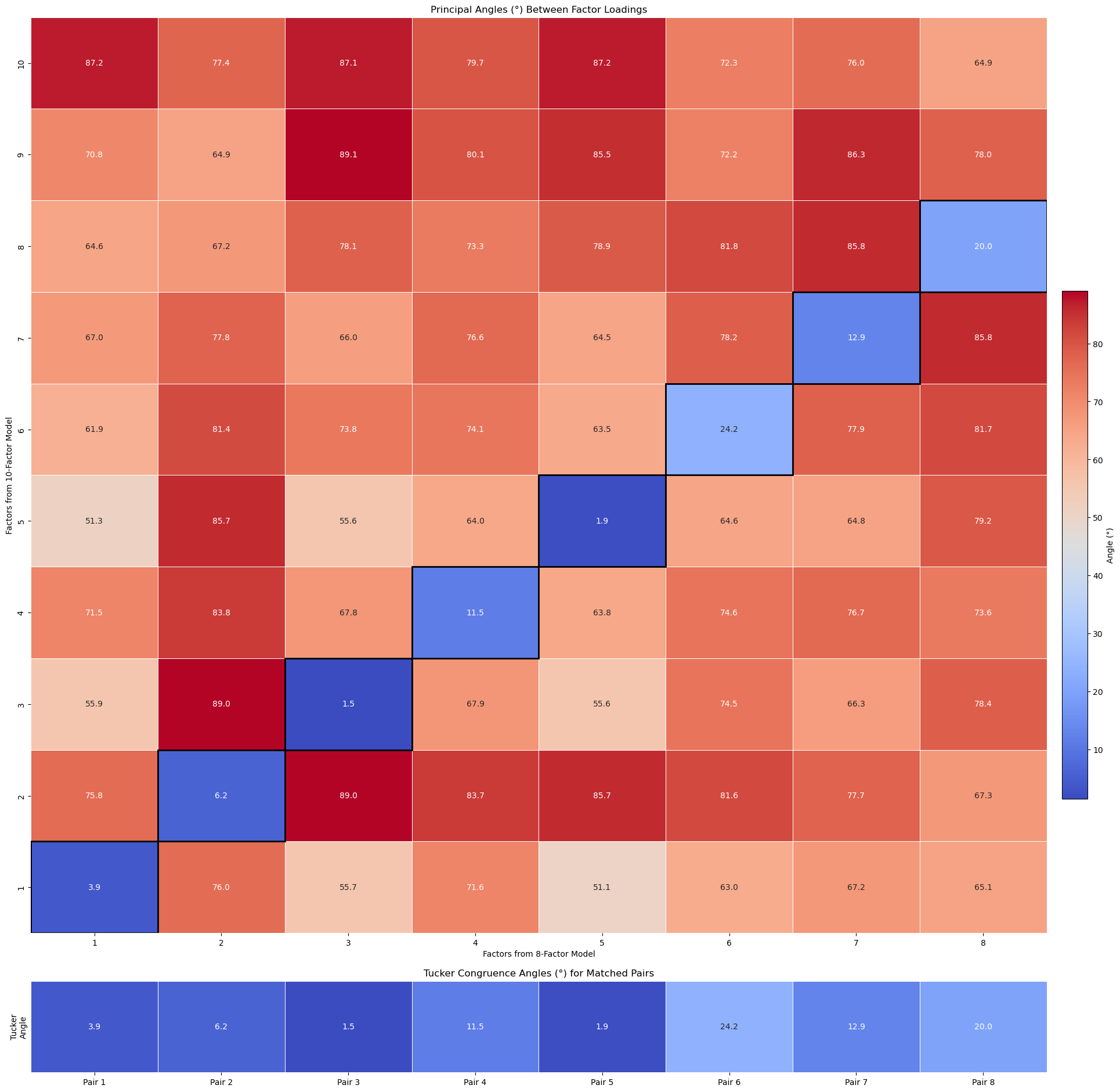}
    \caption{8 vs 10 factors}
    \label{Fig:angle_8_10}
\end{subfigure}

\caption{Principal angles and Tucker's-congruence angles between the 8-factor solution and other factor dimensions}
\label{fig:dimensional_robustness}
\end{figure*}

\paragraph{Robustness to factor dimensionality.}
Fig.~\ref{fig:dimensional_robustness} contrasts the eight-factor reference with models fitted at \(k \pm 1,2\) factors, utilizing principal angles and Tucker's congruence coefficient ($\varphi \ge 0.90$ for strong similarity~\citep{Tucker1951}). 

\paragraph{Robustness to model subsampling.}  \label{subsec:skill_model_robust}

\begin{figure}[t]
\centering
\includegraphics[width=0.48\textwidth]{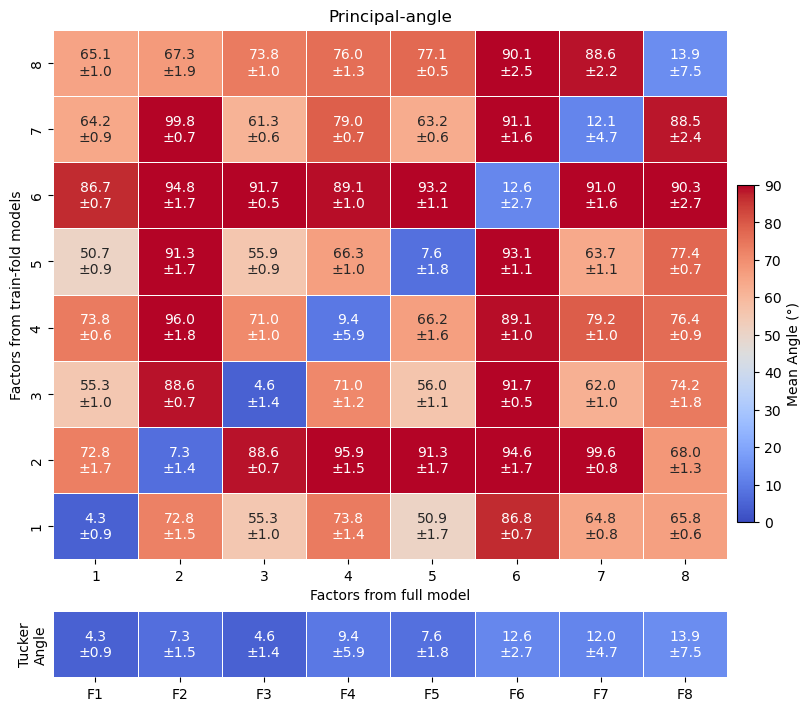}
\caption{The principal angles and Tucker's-congruence angles between the full solution compared to the factors produced by 70\% of the models}
\label{Fig:angle_train_test_model_split}
\end{figure}

Fig.~\ref{Fig:angle_train_test_model_split} shows the mean and std values of the principle angles as well as tucker angles between PAF model produced from 70\% of the models and the full PAF model. 

\begin{figure}[t]
\centering
\includegraphics[width=0.48\textwidth]{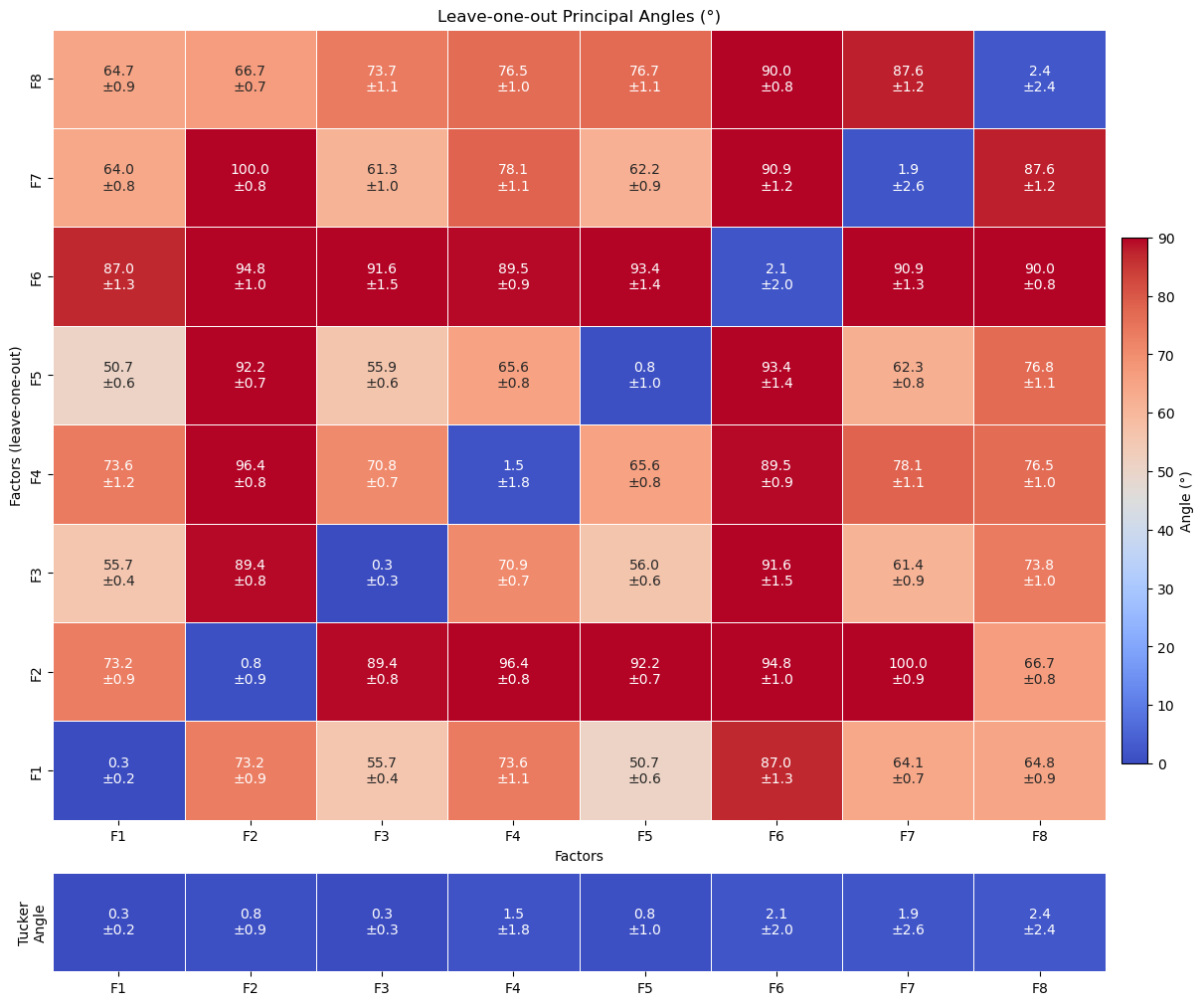}
\caption{The principal angles and Tucker's-congruence angles between the produced factors when a single task is excluded compared to the factors produced by all tasks}
\label{Fig:angle_dataset_split_ours}
\end{figure}

\paragraph{Robustness to task removal.}   \label{subsec:skill_dataset_robust}
Using a leave-one-task-out protocol, we refit the eight-factor PAF on \(B-1\) tasks, aligned the loadings, and computed principal angles (Fig.~\ref{Fig:angle_dataset_split_ours}).

\section{Comparing to chatbot Arena.} \label{subapp:arenastd}
Table~\ref{tab:arena-skill-corr} summarizes for each skill its correlation with Arena score as well as interpretation for why it is.  

\begin{table*}[t]
\centering
\small            
\setlength{\tabcolsep}{4pt}
\resizebox{0.95\textwidth}{!}{
\begin{tabular}{@{}p{2cm}p{1.5cm}p{1.2cm}p{1cm}p{4.2cm}p{4.2cm}@{}}
\toprule
\textbf{Skill} & \textbf{Spearman $\rho$} & \textbf{$p$-value} & \textbf{Case} & \textbf{Interpretation} & \textbf{Top tasks (task $\rho_\text{Arena}$)} \\
\midrule
General NLU & 0.70 & 0.004 & A & Core comprehension and coherence & QQP (.50), MRPC (0.68), WNLI (0.77), IMDB (0.86) \\ 
Entailment \& Bias & \textminus0.81 & <0.001 & D & Over-formality and neutrality reduce perceived helpfulness & QNLI (\textminus.52), RTE (\textminus.83), BBQ (\textminus.76) \\
Long-Doc. Comp. & 0.42 & 0.041 & C & Rewarded in long-context reading; neutral elsewhere & CoQA (.32), XSum (.6), SQuAD (.67), HotpotQA (.97), MS MARCO (.81) \\
Inst. Following / Gen. & 0.25 & 0.130 & C & Helpful in structured prompts; less so in free chat & TruthfulQA-Gen (.24), bAbI (.57), M3Exam(.80), BBQ (\textminus.76) \\
Domain QA & 0.65 & 0.006 & A & Accurate factual recall, domain reasoning & BoolQ (.76), MedQA (.81), OpenBookQA (.82) \\
Social–Ethical Judgment & 0.39 & 0.058 & C & Valued for civility; penalized when overly cautious & CivilComments (.55), MMLU (\textminus.12), BigBench (.47) \\
Prec. \& Fidelity & 0.12 & 0.250 & B & Factual precision overshadowed by fluency & SQuALITY (.36), Math (.62), Quality (.66), GSM8K (.92), BBQ (\textminus.76) \\
Grad-Level Reasoning & 0.29 & 0.120 & B & Deep reasoning under-recognized by raters & GPQA (.24), TriviaQA (.54), Synt-Reason (.47) \\

\bottomrule
\end{tabular}}
\caption{Skill-level correlation with Chatbot Arena Elo. Each skill’s Spearman $\rho$ reflects how its latent-factor scores align with Arena preference rankings. The final column lists representative high-loading tasks (Arena correlation), clarifying which task families drive each latent dimension and how these are valued in Arena evaluations.}
\label{tab:arena-skill-corr}
\end{table*}

\paragraph{Comparing to chatbot Arena.} \label{subapp:arenastd_additional}

\begin{figure}[t]
\centering
\includegraphics[width=0.5\textwidth]{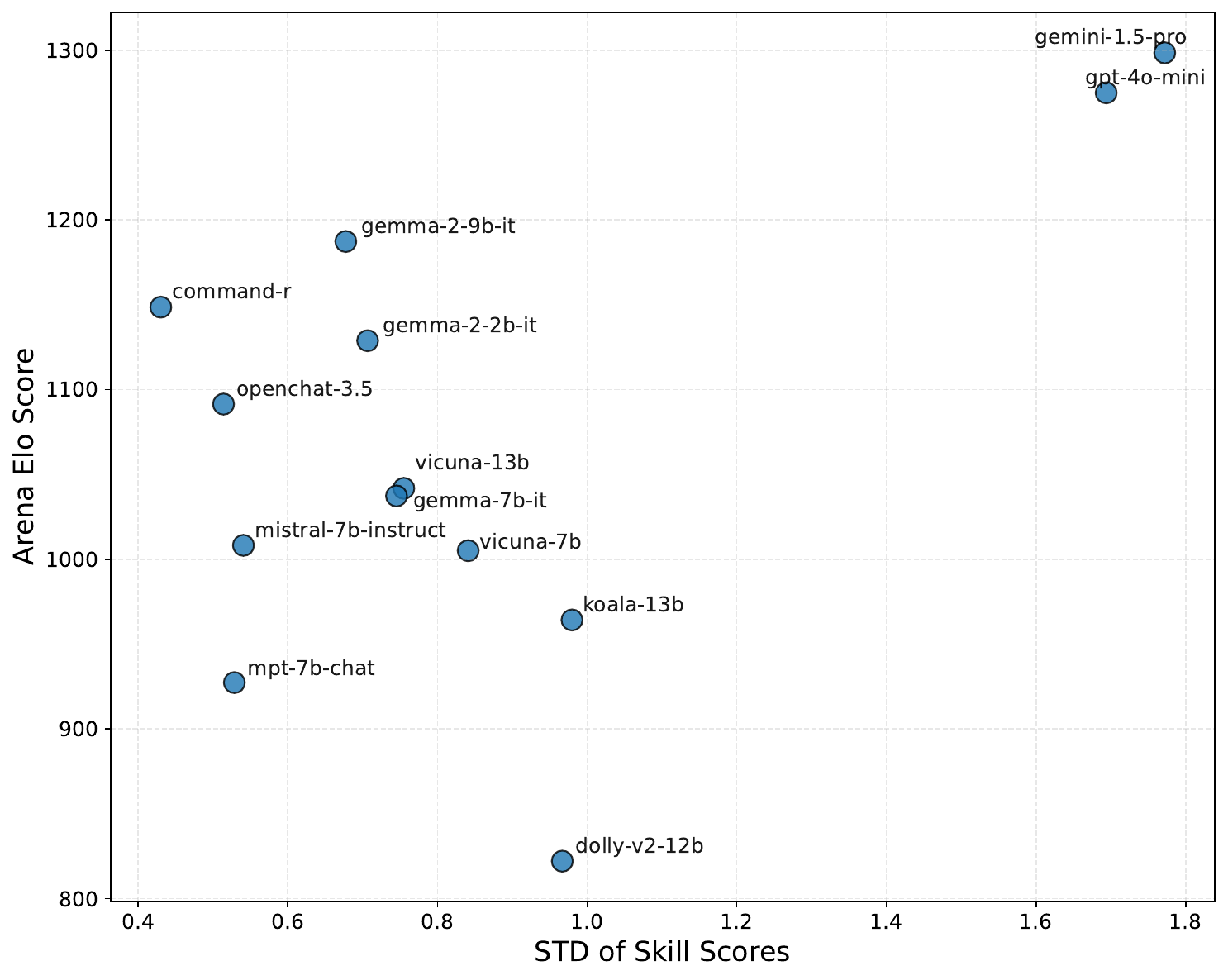}
\caption{Arena ELO score VS skill's STD}
\label{fig:elo_vs_std}
\end{figure}

In Fig.~\ref{fig:elo_vs_std} we plot Arena Elo scores against the across-skill variability (STD) of each model’s eight latent skill scores; each point is a labeled model, with horizontal position indicating how uneven its skill profile is and vertical position its Arena ranking.

\begin{figure}[t]
\centering
\includegraphics[width=0.5\textwidth]{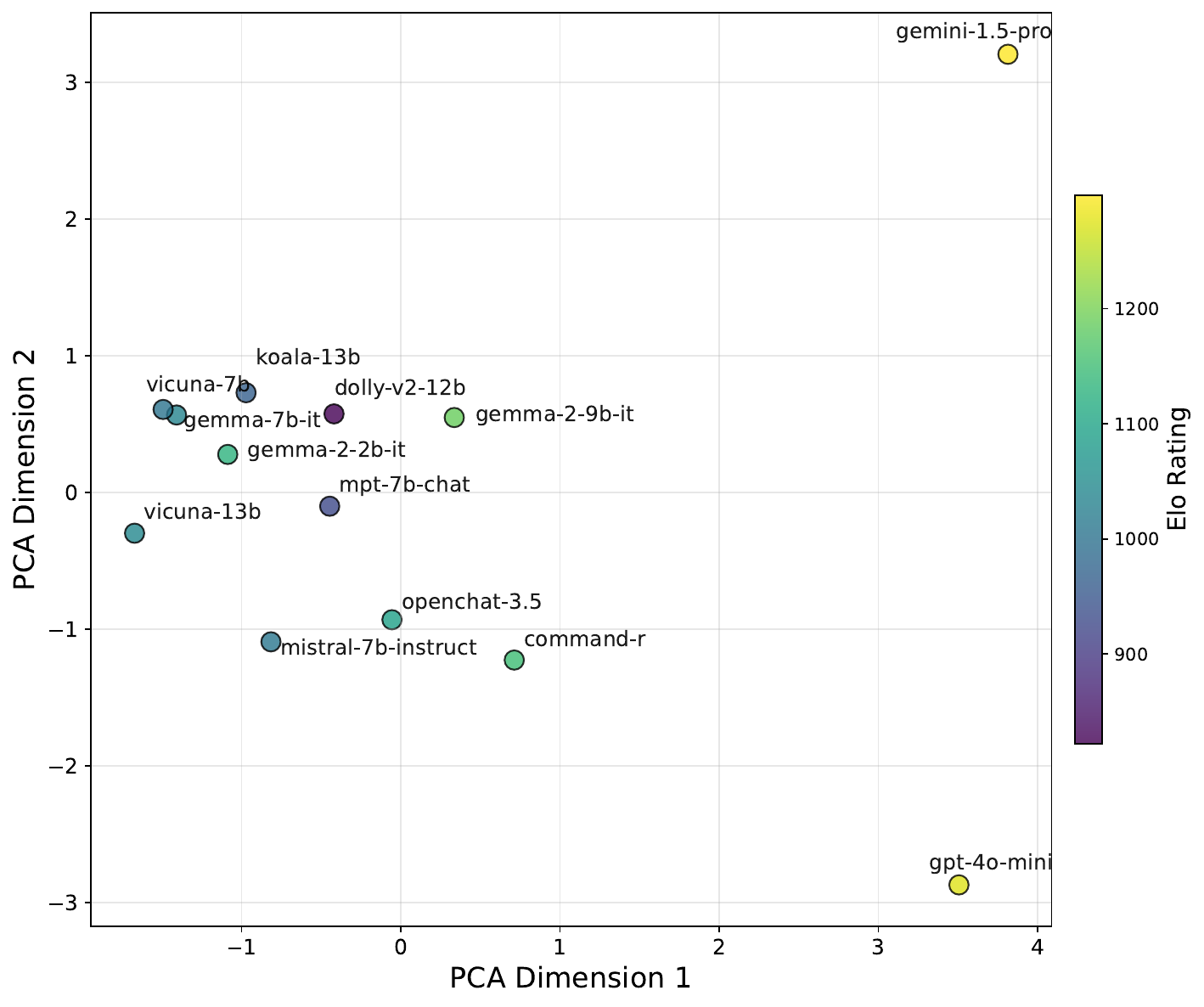}
\caption{Projection of Arena ELO score onto latent skills space}
\label{Fig:arenaProjection}
\end{figure}

In Fig.~\ref{Fig:arenaProjection} we plot she projection of Arena ELO score onto the latent skill space. 

\section{MTEB Results}\label{App:mtebResults}
The \textit{Massive Text Embedding Benchmark} (MTEB) \citep{Muennighoff2023} is a leaderboard for sentence–embedding models.
It evaluates a single model across $>50$ tasks covering semantic textual similarity, retrieval, classification, clustering, and reranking in more than twenty languages.

This section presents the PAF-based identification of latent skills underlying the \textsc{MTEB} leaderboard and similar analysis as shown for our build-in leaderboard.

\subsection{Number of Latent Capabilities}
The optimal number of parameters is \textbf{6~factors} for the \textsc{MTEB} leaderboard.

\subsection{Naming MTEB Skills} \label{App:NamingMteb}

\begin{figure}[t]
\centering
\includegraphics[width=0.5\textwidth]{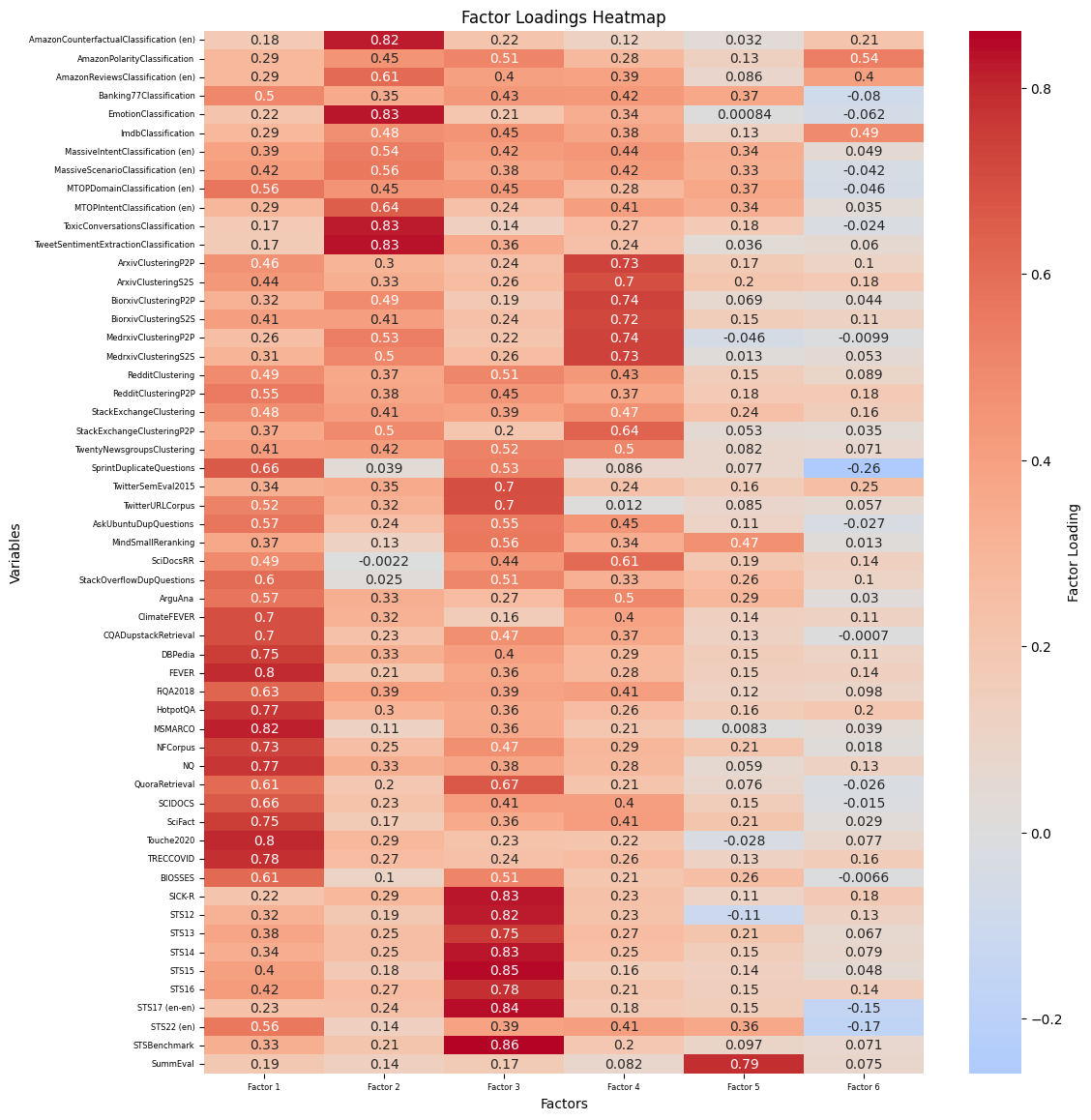}
\caption{Factor Loading for 6 factors based on MTeb leaderboard}
\label{Fig:factor_loading_6_factors}
\end{figure}

Based on the Figure~\ref{Fig:factor_loading_6_factors} the different capabilities are:
\begin{enumerate}
    \item Factual Knowledge \& Information Retrieval: Dominated by tasks involving factual verification (e.g., FEVER), question answering (HotpotQA) and document retrieval (MSMROC). Represents the ability to retrieve, verify, and reason over factual or domain-specific knowledge. 
    
    \item Sentiment \& Intent Understanding: High loadings on tasks requiring nuanced interpretation of emotion detection (e.g., EmotionClassification), sentiment analysis (TonicConversionClassification), and intent recognition (MassiveIntentClassification).
    
    \item Semantic similarity and detection: based on STS12-STS16, STSBenchmark, SICK-R data sets it reflects understanding of textual equivalence, paraphrasing, and similarity scoring.
    
    \item Clustering \& Topic Classification: Strong association with clustering tasks (e.g., topic modeling, semantic grouping). Reflects a model’s ability to identify latent thematic structures in unstructured data based on ArxivClusteringP2, MedrxivClusteringP2 etc.
    
    \item Summarization \& Coherence Evaluation: based on summeval with secondary loadings in tasks requiring coherence assessment.
    
    \item Residual Skills: Most tasks load weakly on Factor 6 (e.g., SummEval: 0.075, Banking7Classification: -0.08, STS22: -0.17), with no clear dominant theme. This suggests it explains minimal shared variance compared to other factors. Some tasks (e.g., STS22, SystemDuplexekQuestions) show small negative correlations with Factor 6. This could indicate minor trade-offs or inverse relationships (e.g., tasks requiring specialized skills vs. general ones), but the lack of strong loadings makes these patterns hard to interpret meaningfully. 
    
    In FA, after extracting major factors, later factors often account for residual covariance or noise. Factor 6 might reflect idiosyncratic variance unique to specific tasks (e.g., SummEval’s focus on summarization evaluation) that doesn’t align with broader skills.

    To validate its significance, we compared the eigenvalue of Factor 6 to those derived from random data. Although it does not fall below the random threshold—indicating that it explains more variance than expected by chance—its weak loadings and lack of thematic coherence continue to pose interpretative challenges.

\end{enumerate}

\subsection{Model Performance Across Skills}
Here, we present factor score $\ThetaMat$ which represent the estimated values of the latent attributes for each model or entity in the task. 
Figure~\ref{fig:factor_scores_ours} displays the leaderboard where each model receive a score summarizing its performance across the different skills. This ranking is particularly useful for identifying the models capabilities on different skills. 

\begin{figure}[t]
\centering
\includegraphics[width=0.4\textwidth]{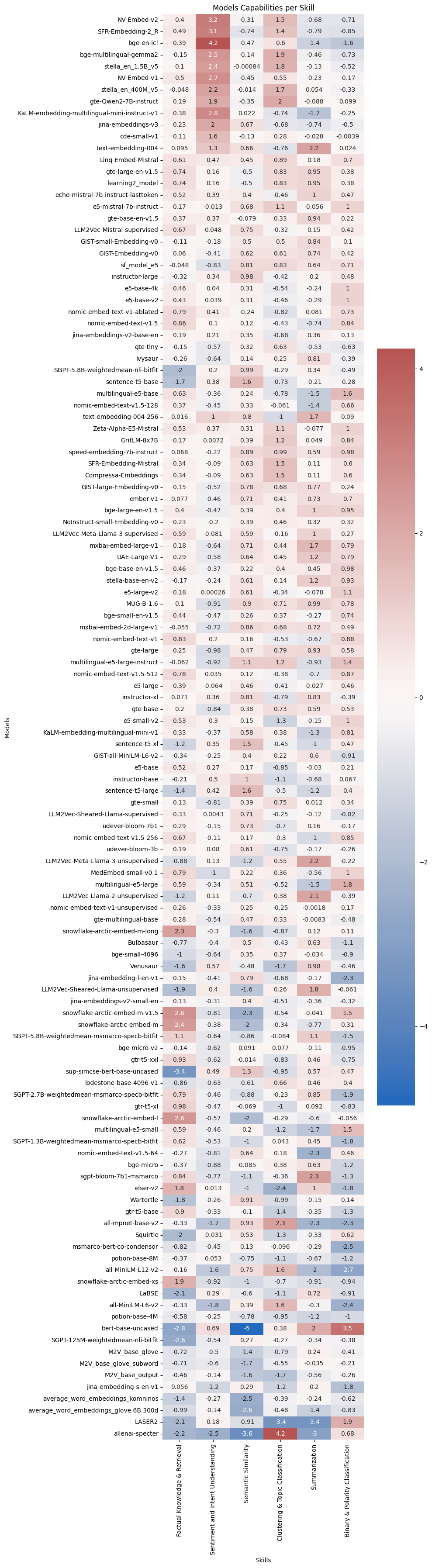}
\caption{Factor Scores for 6 factors based on MTeb leaderboard}
\label{Fig:factor_scores_mteb}
\end{figure}

Figure~\ref{Fig:factor_scores_mteb} displays the leaderboard for MTEB data.

\subsection{Task-Skill Correlations}

Figure~\ref{Fig:factor_loading_6_factors} presents the factor loadings $\LambdaMat$, illustrating how each task relates to the identified latent skills. 
For instance, \texttt{SummEval} exclusively loads onto Factor5, aligning with its specific focus on summarization quality. 
In contrast, \texttt{QuoraRetrieval} is associated primarily with Factors1 and~3, reflecting its intended evaluation of semantic similarity and question retrieval capabilities.

This analysis also reveals unexpected relationships among tasks. 
For example, \texttt{X} and \texttt{Y} both load strongly on Factor2, suggesting a shared underlying capability related to nuanced sentiment interpretation, despite their distinct task definitions and purposes. 
Conversely, tasks traditionally assumed to tap into similar broad categories, such as clustering tasks (\texttt{}) and classification tasks (\texttt{}), load onto distinct factors (Factor4 and Factor~1, respectively), indicating that these superficially related tasks measure fundamentally different latent skills. These insights highlight the importance of empirically validating task relationships rather than relying solely on their superficial or categorical descriptions.

\subsection{Analysis of the Latent Space}

\textbf{Uniqueness} 
All MTEB tasks have uniqueness values below 0.4 \citep{Costello2005}, indicating that the factor model accounts well for their performance patterns.

\subsection{Robustness of Latent Skills}

As before, we analyze the stability of the factor structure under two perturbations: varying the number of factors and removing individual models. In both cases, principal angle analysis quantifies subspace alignment, with lower angles indicating greater stability.

\paragraph{Varying Number of Factors}
For MTEB, where we use 6-factor solution, we compare models with 3, 6, and 10 extracted factors. As shown in the 3-factor model captures a coarse but stable approximation of the 6-factor solution, with moderate angular deviations indicating partial alignment. In contrast, increasing to 10 factors retains strong alignment for the core structure (angles $<5^\circ$ for most shared factors) but introduces additional dimensions with high angular drift (>$70^\circ$), suggesting overfitting or modeling of noise. These results demonstrate that the six-factor solution captures a stable and interpretable core while avoiding the instability introduced by unnecessary complexity.

\paragraph{Robustness to Model Subset Variation}\label{subsec:skill_model_robust_mteb}
We test the sensitivity of the latent space by performing 5-fold cross-validation (splitting the models into training (80\%) and test (20\%) subsets). For each fold, we independently recomputed the PAF model on both subsets and quantified the alignment between factor spaces using principal angle analysis.

Low angles along the diagonal indicate strong agreement between factors learned from different model subsets. The first three factors exhibit high stability, while Factors 5 and 6—show high angular deviation ($>70^\circ$), suggesting they capture more task-specific or fragile structure.

Together, these analyses provide strong evidence that the skills identified by the PAF model represent stable and generalizable LLM capabilities.

\paragraph{Robustness Across Models}
To evaluate whether new models fall within the representational scope of the learned capability space, we test each model for outlier behavior. We computed the Mahalanobis distance \citep{mahalanobis1936generalised} between the model's performance vector $\boldsymbol{\phi}$ and the distribution of existing models. 
In 5-fold MTEB validation, no model exceeded this threshold, suggesting that the existing factor space robustly accommodates current architectures.

\section{Evaluate New Model Skill Profile} \label{app:new_model}

\begin{figure}[t]
\centering
\begin{subfigure}[b]{0.48\linewidth}
    \centering
    \includegraphics[width=\linewidth]{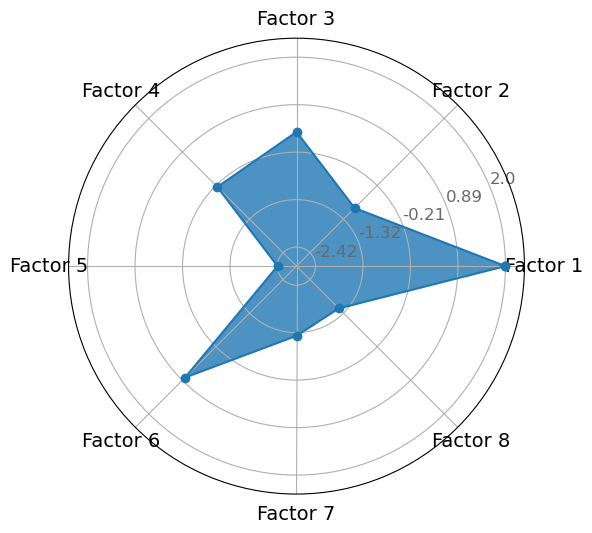}
    \caption{Flan-T5-XL}
    \label{Fig:t5_xl_factors}
\end{subfigure}
\hfill
\begin{subfigure}[b]{0.48\linewidth}
    \centering
    \includegraphics[width=\linewidth]{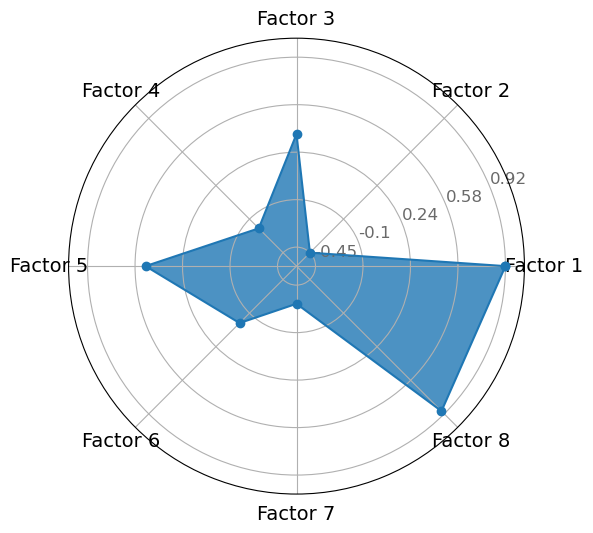}
    \caption{OpenChat-3.5}
    \label{Fig:openchat_factors}
\end{subfigure}
\caption{Projection of new models}
\label{Fig:new_model_projection}
\end{figure}

In Fig.~\ref{Fig:new_model_projection} we show the evaluated skill profile for T5 and openchat models. 

\end{document}